\documentclass{article} 
\usepackage{iclr2027_conference,times}
\iclrfinalcopy

\usepackage{amsmath,amsfonts,amssymb,bm}

\def\eqref#1{equation~\ref{#1}}

\def\1{\bm{1}}

\def\vc{{\bm{c}}}

\def\ve{{\bm{e}}}

\def\vh{{\bm{h}}}

\def\vp{{\bm{p}}}
\def\vq{{\bm{q}}}
\def\vr{{\bm{r}}}

\def\vu{{\bm{u}}}

\def\vx{{\bm{x}}}

\def\mI{{\bm{I}}}

\def\mX{{\bm{X}}}

\def\mZ{{\bm{Z}}}

\DeclareMathAlphabet{\mathsfit}{\encodingdefault}{\sfdefault}{m}{sl}
\SetMathAlphabet{\mathsfit}{bold}{\encodingdefault}{\sfdefault}{bx}{n}

\newcommand{\Ls}{\mathcal{L}}
\newcommand{\R}{\mathbb{R}}

\usepackage{hyperref}
\usepackage{url}
\usepackage{booktabs}
\usepackage{graphicx}
\usepackage{multirow}
\usepackage{xcolor}
\usepackage{subcaption}
\usepackage{wrapfig}
\usepackage{algorithm}
\usepackage{algpseudocode}

\newcommand{\Grass}{\mathrm{Gr}}

\newcommand{\cf}{\textit{cf.}}

\title{Neural Data Needs Semantic Tokenization: \\
Behavioral Events as Boundaries of Session-Transferable Tokens}

\author{{Sangyoon Bae} \\
Interdisciplinary Program in Artificial Intelligence \\
Seoul National University\\
Seoul, 08826, South Korea \\
\texttt{stellasybae@snu.ac.kr} \\
\And
Jiook Cha \\
Department of Psychology \\
Seoul National University\ \\
Seoul, 08826, South Korea  \\
\texttt{connectome@snu.ac.kr}}

\begin{document}

\maketitle

\begin{abstract}
Extracellular electrophysiology records a different set of neurons in every session. Neural foundation models embed each neuron and each session into their tokens, so every new session is an input they have never seen, and they fail to generalize to it. A tokenizer for new sessions needs a unit that every session shares and that carries behavior. Population activity offers such a unit. It evolves on a low-dimensional manifold that persists across neuronal turnover and across animals once sessions are aligned. This manifold changes regime at task events such as stimulus onset and movement onset. Within each regime the population occupies a state, the part of the manifold it spans between two events, and each state carries its own behavioral meaning. We propose \textbf{Tokenization with States} (TWS), which segments each trial, one repetition of the task, at these events and converts every state into tokens of population geometry, with no neuron or session embedding. On held-out International Brain Laboratory (IBL) sessions, TWS decodes movement even from regime boundaries that carry no information about the target, while a per-neuron foundation model pretrained on those sessions decodes at chance. Frozen after training on mice alone, TWS transfers to macaques and Utah arrays with only a linear probe. On reach direction, a target that its boundaries do not define, it achieves a Matthews correlation of $0.23$, where the event time alone achieves $0.01$. For cross-session generalization, the token matters more than the model on top.
\end{abstract}

\section{Introduction}
\label{sec:intro}

\begin{figure}[t]
\centering
\includegraphics[width=0.8\textwidth]{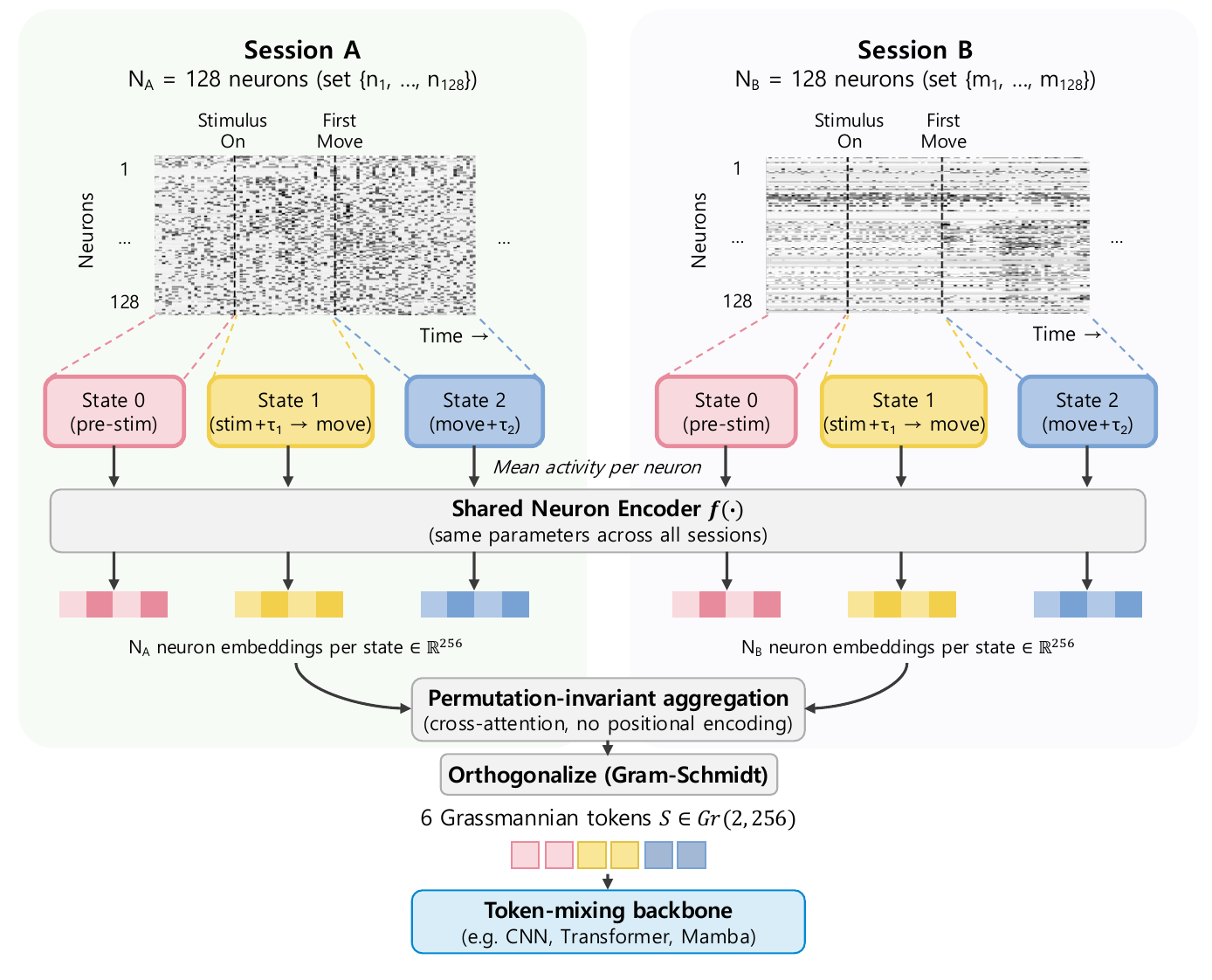}
\caption{\textbf{Overview of Tokenization with States (TWS).} Each session records a different set of neurons. TWS segments every trial into behavioral states at task events and converts each state into tokens without neuron identity, so all sessions share one token space. States S0 to S2 are the states $k{=}1,2,3$ of Algorithm~\ref{alg:tws}, and $\tau_1$ and $\tau_2$ are the event times shifted by two trainable offsets shared by all sessions.}
\label{fig:overview}
\end{figure}

Every extracellular electrophysiology session records a different set of neurons. Electrode positions shift between Neuropixels insertions, impedance changes alter unit isolation with Utah arrays, and spike sorting yields different populations even from identical trajectories~\citep{jun2017fully, dhawale2017automated, steinmetz2021neuropixels}. Every new session is therefore \emph{out-of-distribution} by default. Neural foundation models are exposed to this shift, because every token they build carries the identity of one neuron and of the session it came from. Neuron overlap across sessions is typically zero (Appendix~\ref{app:neuron_overlap}): no neuron label recurs, so per-neuron tokens form a vocabulary that is entirely session-specific and collapses on unseen sessions.

A tokenizer for new sessions therefore needs a unit with two properties: (1) a single decoder can use it in every session, and (2) it carries behavioral meaning. We look for such a unit in population activity, the joint activity of the recorded neurons. It evolves on a low-dimensional manifold~\citep{cunningham2014dimensionality, gallego2017neural}. This manifold separates into distinct regimes at behavioral events such as movement onset~\citep{vyas2020computation, churchland2012neural}.\footnote{These regimes differ from the events of event segmentation theory (Appendix~\ref{app:est_distinction}).} \citet{elsayed2016reorganization} showed that at movement onset population activity moves into a subspace orthogonal to the one it occupied during preparation. (1) The manifold is shared across sessions. The same latent dynamics persist across months of neuronal turnover~\citep{gallego2020long} and across animals~\citep{safaie2023preserved}. These studies align each session to the shared structure with a transform such as CCA, and a unit that describes the manifold without neuron identity needs no such alignment. (2) The manifold carries behavior. Within the window between two behavioral events, it tracks what the animal does~\citep{safaie2023preserved}. The manifold of population activity between two behavioral events therefore meets both requirements, and we use it as the unit of a token (Appendix~\ref{app:extended_related}).

Language modeling has shown that the definition of a token matters. Whole words form a closed vocabulary that cannot represent unseen words~\citep{sennrich2016neural}. Single characters carry little meaning on their own and produce long sequences. Subword tokenizers such as BPE~\citep{sennrich2016neural} and SentencePiece~\citep{kudo2018sentencepiece} occupy an intermediate level of granularity. They learn reusable, semantic units below the level of whole words from the frequency of character sequences. Neural data show both failure modes. Per-neuron tokens form a closed vocabulary, and single time bins, like single characters, carry little meaning on their own.

We therefore propose a new tokenizer that converts each behavioral regime of a trial, one repetition of the task from stimulus to response, into a token of the population manifold: \textbf{Tokenization with States} (TWS; Figure~\ref{fig:overview}), where a state is the population activity of one behavioral regime. Many controlled electrophysiology experiments record the time of each task event of a trial, such as stimulus onset and the first movement. TWS segments the trial at these times and converts the activity of each regime into a few tokens: the subspace that this activity spans, built from all recorded neurons without their order, so the tokens need no neuron or session embedding. A backbone then mixes these tokens into a prediction of behavior. What the tokens carry therefore bounds what any backbone can decode, and for cross-session generalization the tokenizer decides more than the model on top.

\textbf{Contributions.}

\begin{itemize}
    \item \textbf{Tokens built per behavioral regime rather than per neuron prevent decoding from collapsing on sessions with new neurons.} On unseen IBL sessions TWS decodes movement even when its regime boundaries carry no information about the target, while a per-neuron foundation model decodes at chance even after pretraining on those sessions and their movement labels.
    \item \textbf{A frozen TWS trained on mice transfers to other laboratories, to macaques and to Utah arrays.} On choice and reach direction, targets that the token boundaries do not define, a linear probe on it outperforms every baseline of Table~\ref{tab:cross_dataset}, including a per-neuron foundation model trained on the target datasets themselves.
    \item \textbf{A token space shared by every session lets a decoder fit on a few sessions generalize to the others.} On reaction time, a probe fit on five sessions outperforms every baseline of Table~\ref{tab:main} fit on all sessions.
\end{itemize}

\section{Related Work}
\label{sec:related}

\textbf{Neural foundation models.} Existing multi-session models retain neuron or session identity: POYO~\citep{azabou2023unified}, POYO+~\citep{azabou2025multisession} and NDT2~\citep{ye2023neural} learn an embedding for every neuron, session or context, NEDS~\citep{zhang2025neural} finetunes on each new session, and CEBRA~\citep{schneider2023learnable} needs a model or an input layer for every session; FALCON~\citep{karpowicz2024falcon} benchmarks the recalibration this requires. Alignment methods~\citep{gallego2020long, degenhart2020stabilization, karpowicz2022stabilizing} fit a transform to every new session. SPINT~\citep{le2025spint} is invariant to the order of the units but infers an identity for each unit from unlabeled trials of the new session. TWS infers no unit identity and fits no parameter to a new session, and its tokens are the states between task events (Appendix~\ref{app:alignment_vs_tokenization}). Appendix~\ref{app:extended_related} covers further related work on neural manifolds and their behavioral meaning as well as subword tokenization and data-driven segmentation of neural states.

\section{Method}
\label{sec:method}

\subsection{Problem Setup}

Session $s$ records $N_s$ neurons over $T$ time bins per trial and gives trial activity $\mX^{(s)} \in \R^{T \times N_s}$ where both $N_s$ and the identity of the neurons change from session to session. We seek a representation $\vh \in \R^d$ of a trial that captures task structure whichever neurons the session records.

\textbf{Which variables such a representation can carry.} Some behavioral variables such as locomotion and arousal occupy a low-dimensional subspace shared across the population~\citep{stringer2019spontaneous, musall2019single} and any large subset of neurons carries them. Others such as choice~\citep{steinmetz2019distributed} are decoded along directions that differ from session to session. We call the first kind Type~A and the second kind Type~B. A representation without neuron identity is designed to preserve Type~A variables, while Type~B variables need not survive the removal of neuron identity. TWS therefore targets Type~A variables across sessions (Appendix~\ref{app:type_ab_variability}).

\subsection{TWS: Tokenization with States}

TWS segments a variable-neuron trial $\mX \in \R^{T \times N_s}$ along time into $K{=}3$ behavioral regimes and converts each one into $k_\mathrm{dim}{=}2$ tokens of dimension $d{=}256$, six tokens for the trial (Figure~\ref{fig:overview}, Algorithm~\ref{alg:tws}). Soft sigmoid gates place the boundaries at known event times with learnable offsets:
\begin{align}
    w_1(t) &= \sigma\!\left(\frac{b_\mathrm{stim} - t}{\gamma}\right), \quad
    w_2(t) = \sigma\!\left(\frac{t - \tau_1}{\gamma}\right) \cdot \sigma\!\left(\frac{b_\mathrm{move} - t}{\gamma}\right), \quad
    w_3(t) = \sigma\!\left(\frac{t - \tau_2}{\gamma}\right)
    \label{eq:soft_weights}
\end{align}
where $\tau_1 {=} b_\mathrm{stim} {+} \Delta_1$ and $\tau_2 {=} b_\mathrm{move} {+} \Delta_2$, $\Delta_1$ and $\Delta_2$ are two trainable scalars shared by all trials and sessions and bounded by a sigmoid (learned values in Appendix~\ref{app:training}), and $\gamma$ is a temperature. Population activity does not switch regime at the event itself but settles into the new one after a delay, so each boundary carries a trainable offset instead of a fixed one. The two transitions, $[b_\mathrm{stim}, \tau_1)$ and $[b_\mathrm{move}, \tau_2)$, fall outside the three states. Before gating each neuron's spike counts go through $\log(1+x)$ and a z-score over the trials of its session in the batch. This step uses only the unlabeled activity of that session. One encoder lifts the mean activity of every neuron to $\R^d$, and one cross-attention aggregates the $N_s$ vectors with no positional encoding over them. Both are shared by every state, neuron and session; only the $k_\mathrm{dim}$ queries that extract a state differ between states. Each state gives $k_\mathrm{dim}{=}2$ tokens on $\Grass(k_\mathrm{dim}, d)$ (other values in Appendix~\ref{app:ablations}) and four blocks of a kernel-3 convolution mix the six tokens (Appendix~\ref{app:backbone}). Each token set is an orthonormal basis, a point on a Grassmannian manifold~\citep{edelman1998geometry}, which bounds what one state can carry. The bound is what matters, not the geometry: tokens that skip the Gram--Schmidt step decode within $0.01$ of these (Appendix~\ref{app:no_gs}).

\textbf{Property (permutation invariance).} The claim that TWS tokens carry no neuron identity rests on one property of this aggregation. Permuting the $N_s$ recorded neurons together with their coordinates leaves every TWS token unchanged because each step before the tokens either acts on each neuron alone or sums over the neurons with a softmax that has no positional encoding over them (proof and numerical check in Appendix~\ref{app:implementation}). The tokens do depend on how many units a session records: with a quarter of the units TWS retains $83\%$ of its movement decoding (Appendix~\ref{app:global_pool}).

\begin{algorithm}[t]
\caption{Tokenization with States (TWS): forward pass for one trial}
\label{alg:tws}
\begin{algorithmic}[1]
\Require Trial activity $\mX \in \R^{T \times N_s}$, event times $b_\mathrm{stim}, b_\mathrm{move}$ ($b_\mathrm{move}{=}T$ if no first movement falls inside the window)
\Ensure Tokens $\{\ve_{k,j}\} \in \R^{6 \times d}$, $d{=}256$; trial embedding $\vh$ \Comment{$N_s$ neurons $\to$ 6 tokens of dim 256}
\State $\vx(t) \gets$ z-score of $\log(1+\vx(t))$ over the trials of its session in the batch \Comment{per neuron}
\Statex \textcolor{gray}{\textit{Step 1: Regime boundary segmentation (\cf~Fig.~\ref{fig:overview}, state boxes)}}
\For{state $k = 1, \ldots, K$ \textbf{with} $K{=}3$}
    \State Compute soft weights $w_k(t)$ via sigmoid gates at $b_\mathrm{stim}, b_\mathrm{move}$ \Comment{Eq.~\ref{eq:soft_weights}}
    \State $\bar{w}_k(t) \gets w_k(t) \,/\, \textstyle\sum_{t'} w_k(t')$
\EndFor
\Statex \textcolor{gray}{\textit{Step 2: State-weighted population summary (\cf~Fig.~\ref{fig:overview}, shared encoder)}}
\For{state $k = 1, \ldots, K$}
    \State $\bar{\vx}_k \gets \textstyle\sum_{t=1}^{T} \bar{w}_k(t) \cdot \vx(t) \in \R^{N_s}$ \Comment{mean activity per neuron per state}
\EndFor
\Statex \textcolor{gray}{\textit{Step 3: Manifold commitment (\cf~Fig.~\ref{fig:overview}, aggregation and Gram--Schmidt)}}
\For{state $k = 1, \ldots, K$}
    \For{neuron $n = 1, \ldots, N_s$}
        \State $\vu_n \gets f_\mathrm{coord}(\vp_n) + f_\mathrm{act}(\bar{x}_{k,n})$ \Comment{scalar $\to$ $\R^d$; $f_\mathrm{act}$ shared by all states and sessions, $f_\mathrm{coord}$ a shared bias without coordinates (Appendix~\ref{app:implementation})}
    \EndFor
    \For{$j = 1, \ldots, k_\mathrm{dim}$} \Comment{$k_\mathrm{dim}{=}2$}
        \State $\vr_{k,j} \gets \mathrm{LN}\!\left(\mathrm{CrossAttn}(\vq_{k,j},\; \{\vu_n\}_{n=1}^{N_s})\right)$ \Comment{$N_s$ neurons $\to$ 1 vector $\in \R^d$; one $\mathrm{CrossAttn}$ for all states, query $\vq_{k,j}$ per state}
    \EndFor
    \State $\{\ve_{k,j}\}_{j=1}^{k_\mathrm{dim}} \gets \mathrm{GramSchmidt}(\{\vr_{k,j}\})$ \Comment{orthonormal basis $\in \Grass(k_\mathrm{dim}, d)$}
\EndFor
\Statex \hfill \textcolor{gray}{$\triangleright$ \textit{$N_s$ neurons $\to$ $K {\times} k_\mathrm{dim} {=} 6$ tokens of dim $d{=}256$. Invariant to the order of the neurons.}}
\Statex \textcolor{gray}{\textit{Step 4: Backbone + behavioral decode}}
\State $\vh \gets \mathrm{Backbone}\!\left([\ve_{1,1}, \ve_{1,2}, \ve_{2,1}, \ve_{2,2}, \ve_{3,1}, \ve_{3,2}]\right)$ \Comment{$\in \R^{6 \times d}$}
\State $\hat{y} \gets \mathrm{LinearProbe}\!\left(\mathrm{mean}(\vh)\right)$ \Comment{evaluation; training uses one linear head per state}
\end{algorithmic}
\end{algorithm}

\section{Experiments}
\label{sec:experiments}

\subsection{Datasets and Setup}
\label{sec:setup}

\textbf{Data.} \textbf{IBL Brain Wide Map}~\citep{ibl2025brainwide}: 243 Neuropixels sessions from 109 mice in a visual decision task. \textbf{Steinmetz 2019}~\citep{steinmetz2019distributed}: 39 Neuropixels sessions from 10 mice. \textbf{Gallego et al.}~\citep{gallego2020long}: 111 Utah array sessions from 4 reaching macaques; reach direction is defined in the 84 center-out sessions and we score it there. \textbf{NLB MC\_Maze}~\citep{pei2021neural}: 23 Utah array pseudo-sessions from 1 macaque. Each session enters with at most 128 units chosen without labels (most spikes for IBL, highest variance elsewhere, zero-padded when fewer; Appendix~\ref{app:training}), in 20\,ms bins from 0.5\,s before stimulus onset or go cue to 1.5\,s after it ($T{=}100$; for NLB 1.2\,s, $T{=}85$, so that every bin stays inside its trial). The IBL movement label covers $2$\,s after stimulus onset (Appendix~\ref{app:glossary}), so a first movement in the last half second makes the label positive while TWS gets no movement boundary.

\textbf{Model versions.} Four versions of one architecture separate what the architecture, pretraining and behavioral training contribute: \textbf{TWS (random init)}, a forward pass with random weights; \textbf{TWS pretrained}, after self-supervised training that reconstructs masked state means and pushes consecutive state subspaces toward orthogonality (Appendix~\ref{app:training}); \textbf{TWS from scratch}, trained on behavior alone; and \textbf{TWS pretrained + finetuned}. Every IBL model is scored on 53 held-out sessions (49 drawn at random and the 4 test sessions of NEDS~\citep{zhang2025neural}). TWS and PerceiverIO train without them. The baselines that fit a model to each session fit one to every held-out session by design.

\textbf{Evaluation.} We decode seven IBL variables (Appendix~\ref{app:glossary}, \ref{app:class_dist}); the main text reports movement, reaction time (RT), choice and block, and Appendix~\ref{app:full_results} all seven and the within-session scores. Cross-session scores are leave-one-session-out: a Ridge probe on frozen embeddings, fit on the other sessions, with no labels and no fitted parameters from the test session. Its unlabeled trials set the per-neuron z-score, and continuous targets are standardized within each session, which uses the label statistics of the test session. We report the Matthews correlation coefficient (MCC; $0$ at chance) for discrete variables, $R^2$ for continuous ones, and \emph{session identifiability} $\mathrm{MI}(\vh;\,\mathrm{session})/H(\mathrm{session})$.

\textbf{Baselines.} POYO~\citep{azabou2023unified} stands for per-neuron models; it pretrained on all IBL sessions, the held-out ones included, with an auxiliary loss on movement, wheel velocity and licking. Supervised CEBRA~\citep{schneider2023learnable} and NEDS~\citep{zhang2025neural} fit one model per session, NEDS with its own per-neuron and per-session embeddings. PerceiverIO~\citep{jaegle2021perceiver} is the POYO tokenizer without neuron or session embeddings on the TWS CNN backbone, trained from scratch. Aligned PCA rotates each session's PCA latents onto a shared template with that session's unlabeled trials (Appendix~\ref{app:alignment_vs_tokenization}), and PCA without alignment serves as a floor. Population Statistics summarizes each trial by the mean and spread of its activity, and Event-time information is the same probe on the first-movement bin alone, with no neural data.

\textbf{Cross-dataset evaluation.} On Steinmetz, Gallego et al.\ and NLB, TWS and PerceiverIO retain their IBL weights and only a Ridge classifier is fit, on all sessions of the dataset but the test one. CEBRA fits one model per session and POYO trains on each dataset (Appendix~\ref{app:xds_finetune}). NLB pseudo-sessions share neurons, so NLB tests transfer across trials only.

\subsection{TWS Tokens Decode Unseen Sessions Where Per-Neuron Tokens Collapse}
\label{sec:main_results}

\begin{table}[t]
\centering
\caption{Cross-session decoding on IBL over the 53 held-out sessions (Section~\ref{sec:setup}). $\pm$: SD over 5 model seeds. PCA and Population Statistics have no trained model; CEBRA and NEDS train one model per session and are run once.}
\label{tab:main}
\resizebox{\textwidth}{!}{
\begin{tabular}{lcccc}
\toprule
 & \multicolumn{2}{c}{Type~A} & \multicolumn{2}{c}{Type~B} \\
\cmidrule(lr){2-3}\cmidrule(lr){4-5}
Method & Movement & RT ($R^2$) & Choice & Block \\
\midrule
\multicolumn{5}{l}{\textit{Cross-session (leave-one-session-out)}} \\
TWS pretrained + finetuned & $0.553 \pm 0.006$ & $0.398 \pm 0.004$ & $\mathbf{0.040 \pm 0.006}$ & $\mathbf{0.031 \pm 0.003}$ \\
\textbf{TWS from scratch} & $\mathbf{0.565 \pm 0.006}$ & $\mathbf{0.402 \pm 0.003}$ & $0.038 \pm 0.006$ & $0.026 \pm 0.004$ \\
TWS pretrained & $0.428 \pm 0.043$ & $0.247 \pm 0.036$ & $0.022 \pm 0.001$ & $0.019 \pm 0.002$ \\
TWS (random init) & $0.274 \pm 0.018$ & $0.083 \pm 0.011$ & $-0.000 \pm 0.004$ & $0.005 \pm 0.005$ \\
PerceiverIO & $0.303 \pm 0.009$ & $0.169 \pm 0.003$ & $0.014 \pm 0.009$ & $0.011 \pm 0.006$ \\
PCA (aligned) & $0.349$ & $0.226$ & $0.000$ & $-0.005$ \\
PCA (unaligned) & $0.002$ & $0.003$ & $0.027$ & $0.007$ \\
POYO & $0.000 \pm 0.001$ & $-0.001 \pm 0.001$ & $-0.000 \pm 0.002$ & $0.000 \pm 0.002$ \\
NEDS & $0.000$ & $-0.010$ & $-0.022$ & $0.076$ \\
CEBRA & $0.000$ & $0.000$ & $0.005$ & $-0.012$ \\
Population Statistics & $0.000$ & $0.009$ & $0.005$ & $-0.002$ \\
\bottomrule
\end{tabular}
}

\end{table}

\textbf{TWS decodes a session it has never seen.} TWS leads every baseline on movement and on reaction time (Table~\ref{tab:main}), and four baselines enter with an advantage. POYO pretrained on these very sessions and still predicts the same class for every trial. Finetuning POYO end-to-end on IBL behavior does not change this (Appendix~\ref{app:alignment_vs_tokenization}). CEBRA trains one model per session and also predicts a constant for every trial, on the classes and on the continuous targets alike (Appendix~\ref{app:cebra_regression}). Aligned PCA rotates each new session onto a shared template and achieves $62\%$ of the movement decoding of TWS and $56\%$ of its reaction-time decoding. TWS fits nothing per session. With regime boundaries that carry no information about the target, TWS still decodes movement, where POYO, NEDS and CEBRA decode nothing (Section~\ref{sec:ablation_k}). On the five held-out sessions that also record video, TWS decodes whisker motion, a variable that training never uses, twice as well as PerceiverIO, and no model decodes pupil diameter across sessions (Appendix~\ref{app:physio}).

\textbf{Choice and block stay near chance across sessions.} Every method, TWS included, stays near chance on choice and block across sessions (Table~\ref{tab:main}), as Section~\ref{sec:method} predicts for Type~B variables (Appendix~\ref{app:type_ab_variability}). Models that report these variables above chance on IBL, such as NEDS~\citep{zhang2025neural}, finetune on each test session; our cross-session probe fits nothing to the test session. The NEDS block score is the largest number in the two Type~B columns and is still chance: its median over the 53 sessions is $0.000$ and $11\%$ of sessions score above zero ($p{=}0.08$, one-sided Wilcoxon signed-rank test).

\subsection{Regime Boundaries Decide What the Tokens Can Hold}
\label{sec:ablation_k}

Table~\ref{tab:boundary_ablation} shows that what the tokens hold depends on the number of regime boundaries and on whether each boundary is aligned with the recorded event time. Moving every boundary by $100$\,ms reduces movement decoding by $7\%$ and moving it by $200$\,ms by $9\%$, while boundaries that ignore the trial reduce it by $62\%$ when every trial is segmented at one constant bin and by $70\%$ when each trial is segmented at a random bin. Training does not repair a wrong segmentation: with those boundaries $40$ epochs of behavioral training stay below TWS (random init) with the regime boundaries (Table~\ref{tab:main}). Transition tokens between consecutive states change neither number, since the three state subspaces are already nearly orthogonal (Appendix~\ref{app:state_manifold}); Appendix~\ref{app:gate_decomposition} explains why the trial is not segmented more finely.

\textbf{Neural activity and the event time.} TWS decodes movement from neural activity, not only from the event time it receives. With boundaries that carry no information about the target (a fixed, random or shuffled bin), the event time decodes nothing, and TWS retains a third of its movement decoding. The event time also defines movement and constrains reaction time, and a probe on it alone achieves movement MCC $0.91$ and RT $R^2$ $0.44$ (Table~\ref{tab:boundary_ablation}, Event-time information). A large part of this movement score comes from the $11\%$ of held-out trials in which the mouse moves shortly before the stimulus. The label counts these trials as trials without a movement, and the event time identifies them because it falls before the stimulus. Without these trials, TWS achieves $93\%$ of the movement information in the event time, and PerceiverIO, which receives no event time, achieves less than half (Appendix~\ref{app:gate_decomposition}).

\begin{table}[t]
\caption{Tokenization boundary ablation (IBL, 53 held-out sessions). Every row uses the same architecture and only the boundaries change. \textbf{(a)} Number of boundaries. \textbf{(b)} Boundary placement at fixed $K{=}3$. Jittered: true boundary plus Gaussian noise with standard deviation $\sigma$ bins. Fixed: boundary at a constant bin for all trials. Random: uniform draw per trial. Shuffled: real boundaries reassigned across (or within) sessions. Jitter and shuffling move only the boundaries that exist: a trial whose first movement falls outside the window keeps none, as in the true condition. Event-time information: the same probe on the one-hot first-movement bin that each condition gives, with no neural data; it measures how much of the label that input holds (mean over 5 draws of the perturbation). Cross-session leave-one-session-out, as in Table~\ref{tab:main}. $\pm$: SD over 5 model seeds, each with its own boundary randomization.}
\label{tab:boundary_ablation}
\vspace{12pt}
\centering
\small
\setlength{\tabcolsep}{6pt}
\renewcommand{\arraystretch}{0.92}
\resizebox{\textwidth}{!}{
\begin{tabular}{lcccc}
\toprule
 & \multicolumn{2}{c}{TWS} & \multicolumn{2}{c}{Event-time information} \\
\cmidrule(lr){2-3}\cmidrule(lr){4-5}
Boundary condition & Movement & RT $R^2$ & Movement & RT $R^2$ \\
\midrule
\multicolumn{5}{l}{\textit{(a) How many boundaries ($K$)}} \\
$K{=}2$: stimulus onset only & $0.214 \pm 0.007$ & $0.096 \pm 0.007$ & -- & -- \\
$\boldsymbol{K{=}3}$\textbf{: stimulus + first movement} & $0.565 \pm 0.006$ & $0.402 \pm 0.003$ & $0.908$ & $0.443$ \\
$K{=}3$ + transitions between states & $0.563 \pm 0.004$ & $0.403 \pm 0.004$ & $0.908$ & $0.443$ \\
\midrule
\multicolumn{5}{l}{\textit{(b) Where to place the boundary ($K{=}3$ fixed)}} \\
\textbf{True behavioral} & $\mathbf{0.565 \pm 0.006}$ & $\mathbf{0.402 \pm 0.003}$ & $0.908$ & $0.443$ \\
Jittered ($\sigma{=}5$ bins) & $0.523 \pm 0.008$ & $0.399 \pm 0.003$ & $0.712$ & $0.441$ \\
Jittered ($\sigma{=}10$ bins) & $0.515 \pm 0.011$ & $0.392 \pm 0.003$ & $0.577$ & $0.436$ \\
Fixed center (bin 50) & $0.215 \pm 0.005$ & $0.128 \pm 0.003$ & $0.000$ & $0.000$ \\
Fixed late (bin 75) & $0.188 \pm 0.003$ & $0.097 \pm 0.006$ & $0.000$ & $0.000$ \\
Random uniform & $0.171 \pm 0.011$ & $0.095 \pm 0.001$ & $0.000$ & $0.000$ \\
Shuffled across sessions & $0.199 \pm 0.007$ & $0.115 \pm 0.004$ & $0.000$ & $0.000$ \\
Shuffled within session & $0.196 \pm 0.009$ & $0.118 \pm 0.003$ & $0.000$ & $0.000$ \\
\bottomrule
\end{tabular}
}
\renewcommand{\arraystretch}{1.0}

\end{table}

\subsection{The Tokenizer, Not the Backbone, Determines Cross-Session Decoding on IBL}
\label{sec:backbone_recovery}

\begin{table}[t]
\centering
\caption{Tokenizer $\times$ backbone on IBL (53 held-out sessions, cross-session protocol of Table~\ref{tab:main}); every model learns from scratch. $\pm$: SD over 5 model seeds. The CNN row repeats Table~\ref{tab:main}.}
\label{tab:backbone_ablation}
\small
\setlength{\tabcolsep}{4pt}
\begin{tabular}{llcccc}
\toprule
 & & \multicolumn{2}{c}{TWS} & \multicolumn{2}{c}{PerceiverIO} \\
\cmidrule(lr){3-4}\cmidrule(lr){5-6}
 & Backbone & Movement MCC & RT $R^2$ & Movement MCC & RT $R^2$ \\
\midrule
\multirow{4}{*}{\shortstack[l]{Token\\mixing}} & CNN & $\mathbf{0.565 \pm 0.006}$ & $\mathbf{0.402 \pm 0.003}$ & $0.303 \pm 0.009$ & $0.169 \pm 0.003$ \\
 & Mamba-1 & $0.536 \pm 0.032$ & $0.385 \pm 0.035$ & $0.287 \pm 0.015$ & $0.158 \pm 0.005$ \\
 & Mamba-2 & $0.555 \pm 0.009$ & $0.400 \pm 0.005$ & $0.294 \pm 0.012$ & $0.162 \pm 0.012$ \\
 & Transformer & $0.552 \pm 0.006$ & $0.398 \pm 0.002$ & $\mathbf{0.313 \pm 0.008}$ & $0.168 \pm 0.004$ \\
\midrule
\multirow{2}{*}{\shortstack[l]{No\\mixing}} & MLP & $0.459 \pm 0.016$ & $0.305 \pm 0.013$ & $0.290 \pm 0.010$ & $0.166 \pm 0.005$ \\
 & Pooling & $0.465 \pm 0.007$ & $0.288 \pm 0.011$ & $0.291 \pm 0.004$ & $0.167 \pm 0.002$ \\
\bottomrule
\end{tabular}

\end{table}

We separate the effect of the tokenizer from the effect of the backbone. Two tokenizers, TWS and PerceiverIO~\citep{jaegle2021perceiver} (the POYO tokenizer without neuron or session embeddings), meet six backbones: pooling, MLP, CNN, Transformer~\citep{vaswani2017attention}, Mamba-1~\citep{gu2024mamba} and Mamba-2~\citep{dao2024transformers}. Every combination trains from scratch, because PerceiverIO has no self-supervised objective (Table~\ref{tab:backbone_ablation}).

\begin{figure}[t]
\centering
\includegraphics[width=\textwidth]{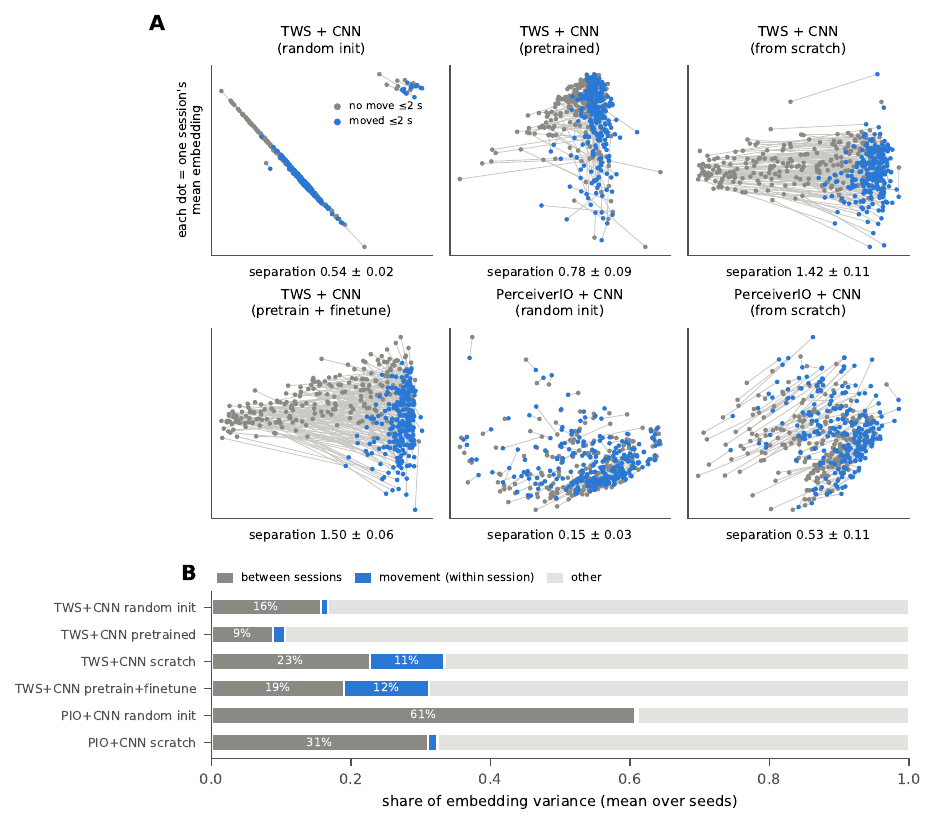}
\caption{\textbf{Training organizes the TWS embedding space around behavior; the tokenizer decides how far.} \textbf{(A)} Each IBL session contributes two points, the mean embedding of its trials with (blue) and without (gray) a first movement within 2\,s, and a line joins the two; PCA of these $486$ points (first seed). Separation: distance between the two behavior centres over the spread of sessions around them, mean $\pm$ SD over 5 seeds. \textbf{(B)} Share of trial-embedding variance explained by session identity and, after removing each session's mean, by movement (mean over seeds). Models trained without the 53 held-out sessions; embeddings of all 243 sessions.}
\label{fig:representation}
\end{figure}

Backbones that mix the tokens differ from one another by at most $5\%$ in movement decoding, while removing the mixing reduces it by $18\%$; their transfer to the other datasets is in Appendix~\ref{app:backbone_sweep}. Swapping TWS for PerceiverIO halves decoding with the same backbone. The embeddings of the two tokenizers also differ in what they spend their variance on. Session identity takes $31\%$ of the PerceiverIO embedding and movement $1\%$, against $23\%$ and $11\%$ for TWS (Figure~\ref{fig:representation}B; Appendix~\ref{app:rate_reduction}). The backbone is interchangeable; the tokenizer is not.

\subsection{Tokens Carry Neural State across Species and Recording Technologies}
\label{sec:cross_dataset}

\begin{table}[!ht]
\centering
\caption{Transfer of the frozen representation with a linear probe; POYO trains on each dataset (protocol in Section~\ref{sec:setup}). MCC, mean $\pm$ SD over 5 seeds. Rows without $\pm$ are deterministic features. Event-time information uses the first-movement bin and no neural data.}
\label{tab:cross_dataset}
\small
\setlength{\tabcolsep}{4pt}
\resizebox{\textwidth}{!}{
\begin{tabular}{lccccc}
\toprule
 & \multicolumn{3}{c}{Steinmetz 2019} & Gallego et al. & NLB MC\_Maze \\
\cmidrule(lr){2-4}\cmidrule(lr){5-5}\cmidrule(lr){6-6}
Method & Choice & Feedback & Movement & Reach direction & Movement \\
\midrule
TWS from scratch & $\mathbf{0.151 \pm 0.015}$ & $\mathbf{0.188 \pm 0.008}$ & $\mathbf{0.581 \pm 0.003}$ & $\mathbf{0.232 \pm 0.010}$ & $\mathbf{0.392 \pm 0.010}$ \\
TWS (random init) & $0.036 \pm 0.016$ & $0.077 \pm 0.017$ & $0.550 \pm 0.007$ & $0.128 \pm 0.002$ & $0.247 \pm 0.010$ \\
CEBRA & $-0.000 \pm 0.000$ & $0.000 \pm 0.000$ & $0.005 \pm 0.050$ & $0.000 \pm 0.010$ & $-0.003 \pm 0.018$ \\
POYO & $0.008 \pm 0.010$ & $0.045 \pm 0.021$ & $0.111 \pm 0.013$ & $0.013 \pm 0.009$ & $0.162 \pm 0.047$ \\
Population Statistics & $-0.007$ & $0.000$ & $0.242$ & $0.046$ & $0.177$ \\
PCA (unaligned) & $0.011 \pm 0.010$ & $0.012 \pm 0.009$ & $0.390 \pm 0.003$ & $0.090 \pm 0.004$ & $0.144 \pm 0.011$ \\
\midrule
\textit{Event-time information} & \textit{$-0.009$} & \textit{$0.149$} & \textit{$0.319$} & \textit{$0.010$} & \textit{$0.260$} \\
\bottomrule
\end{tabular}
}

\end{table}

\textbf{A mouse-Neuropixels-trained TWS transfers to macaques and Utah arrays.} We freeze the TWS trained on IBL and fit only a linear probe on each new dataset. It decodes every target better than every baseline of Table~\ref{tab:cross_dataset}, including POYO, which trains on the target dataset itself.  Training on IBL raises choice, reach direction and NLB movement above TWS (random init), also when the untrained encoder receives biases of the trained size. A PerceiverIO trained on IBL, which also has no neuron or session embedding, transfers too (Appendix~\ref{app:transfer_controls}). The transfer of TWS comes from neural activity: the first-movement time alone decodes neither choice nor reach direction (row Event-time information), and choice holds within each movement group and on go trials alone (Appendix~\ref{app:boundary_only_xds}).

\textbf{What the tokens carry.} Each token is a learned summary of how activity is distributed across the neurons of one regime, and this learned summary is what transfers. Two fixed summaries of the same distribution test this: its mean, and its mean with the spread and quantiles over neurons (Appendix~\ref{app:transfer_controls}). Both approach TWS on movement. On choice and reach direction the better of the two achieves a fifth and a third of TWS. The learned summary needs no time course inside a regime. TWS averages that time course away, and pretraining on data whose temporal order is destroyed transfers at least as well (Appendix~\ref{app:transfer_controls}).

\subsection{The Architecture Sets the Token Space, Behavioral Training Regroups It, Pretraining Lowers Session Identity}
\label{sec:tradeoff}

We compare the four versions of TWS on IBL, defined in Section~\ref{sec:setup}; Appendix~\ref{app:global_pool} adds two controls, one on the head that decodes the tokens and one on the number of units per session.

As shown in Table~\ref{tab:main}, TWS (random init) decodes movement close to a PerceiverIO trained from scratch. The architecture alone already makes movement decodable across sessions. Pretraining lowers the session identity of the embedding, from $23\%$ of the variance after behavioral training alone to $19\%$ when pretraining comes first (Figure~\ref{fig:representation}B), and leaves the final score unchanged (Table~\ref{tab:main}). An earlier backbone shows the same for every combination of its losses (Appendix~\ref{app:loss_ablation}).

In TWS (random init) the trials of each session cluster together whatever their movement state (Figure~\ref{fig:representation}A). Pretraining separates the two movement states only slightly, from $0.54$ to $0.78$ in the separation of Figure~\ref{fig:representation}A, and behavioral training pushes it to $1.50$: the sessions then fall clearly onto two clouds, one per movement state. Subtracting the mean embedding of each session leaves movement decoding intact, trained or not (Appendix~\ref{app:residual_decoding}).

\subsection{Behavior and Session Identity Occupy Different Directions}
\label{sec:invariance}

\begin{figure}[t]
\centering
\includegraphics[width=\textwidth]{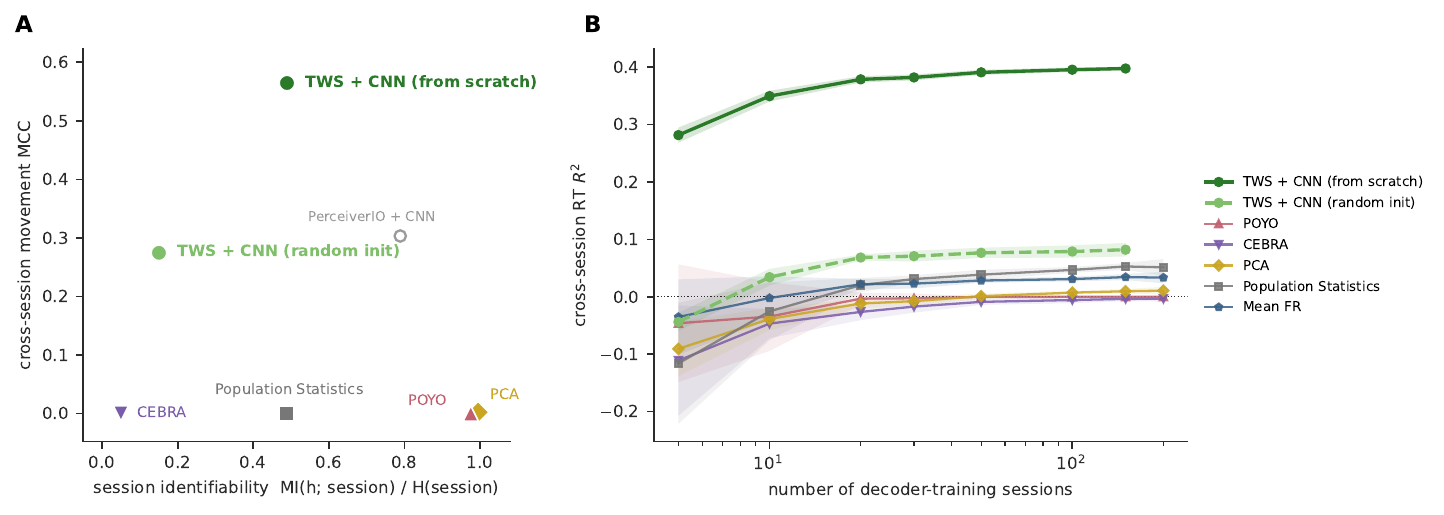}
\caption{\textbf{Session invariance and data efficiency.} \textbf{(A)} Session identifiability (normalized MI of a 243-class probe, $x$-axis) versus cross-session movement MCC ($y$-axis, the values of Table~\ref{tab:main}); methods in the upper-left corner achieve invariance without sacrificing decoding. We draw no error bars because the SD over 5 model seeds is smaller than the markers (values in Appendix~\ref{app:session_stats}). \textbf{(B)} Reaction time $R^2$ versus number of decoder-training sessions (log scale). Shaded bands: ${\pm}1$ SD over 5 model seeds for TWS and over 20 draws of decoder-training sessions for the baselines. Session identifiability over all 243 IBL sessions; TWS and PerceiverIO decoding over the 53 held-out sessions, with decoder-training sessions in (B) drawn from the other 190.}
\label{fig:structural_invariance}
\end{figure}

In TWS, session identity and behavior occupy different directions of the embedding. Deleting the directions that separate sessions leaves movement decoding almost unchanged ($0.56$ against $0.55$). Identifying the session from the two behavior decoding directions works no better than from two random directions (Appendix~\ref{app:session_readout}). The session information that remains therefore does not interfere with decoding. The other methods fail in two ways (Figure~\ref{fig:structural_invariance}A). POYO and PCA identify the session of almost every trial and decode movement at chance, and CEBRA reduces session identity and loses behavior with it. TWS retains behavior with half the session information of POYO and less than PerceiverIO with the same backbone, at the laboratory level as well as the session level (Appendices~\ref{app:info_decomp}, \ref{app:site_invariance} and \ref{app:tsne}). On Steinmetz, a dataset it never saw, the frozen TWS retains a third of the session information of per-neuron features (Appendix~\ref{app:steinmetz_mi}).

\subsection{With Labels from Five Sessions, TWS Outperforms Baselines Trained on All of Them}
\label{sec:data_efficiency}

A decoder for a new laboratory has to be fit on sessions with behavioral labels, and TWS needs few of them. Figure~\ref{fig:structural_invariance}B shows reaction-time decoding against the number of sessions the probe is fit on. With five sessions, TWS from scratch decodes reaction time five times better than the best baseline of Figure~\ref{fig:structural_invariance}B fit on every session. It also exceeds aligned PCA fit on every session by $25\%$ (Table~\ref{tab:main}), and PerceiverIO on the same five sessions fourfold (Appendix~\ref{app:behavior_axis}). Across six models, this score follows how consistently the probe direction points across sessions, not session information (Appendix~\ref{app:behavior_axis}).

\section{Discussion}
\label{sec:discussion}

We propose TWS, a tokenizer that turns the population activity between task events into tokens without neuron or session embeddings. On unseen sessions TWS decodes where per-neuron and per-session models decode at chance. Frozen after training on mice, TWS transfers to macaques and Utah arrays with a linear probe fitted on each dataset, including reach direction, a target that its boundaries do not define. The design sets the limits, and they are the price of the invariance. A tokenizer without neuron identity need not preserve a variable whose decoding direction differs from session to session. Averaging inside a regime gives up the time course within it, and TWS needs the event times that a task records. Free behavior therefore requires a tokenizer that finds the regimes in the data itself. TWS is a practical model for the electrophysiology datasets most laboratories hold. It fits nothing per session, it asks only for the event times a task already records, and a probe fit on a few sessions with behavioral labels generalizes to the remaining sessions.

\newpage
\subsection*{Reproducibility Statement}

All datasets are public and listed with their preprocessing in Section~\ref{sec:experiments}. The tokenizer is specified in Section~\ref{sec:method} and Algorithm~\ref{alg:tws}, the backbone in Appendix~\ref{app:backbone}, and remaining implementation details in Appendix~\ref{app:implementation}. Training procedures and all hyperparameters are in Appendix~\ref{app:training}, and the new-dataset protocol in Appendix~\ref{app:xds_finetune}. Evaluation protocols, baselines and seeds are in Section~\ref{sec:experiments}, and the behavioral variables are defined in Appendix~\ref{app:glossary}. Code will be released upon acceptance.

\subsection*{AI Use Statement}

Generative AI tools (Claude) were used for code assistance during implementation and for drafting portions of this manuscript. All AI-generated content was reviewed, verified, and edited by the authors. We take full responsibility for the final content.

\bibliography{iclr2027_conference}

@article{dhawale2017automated,
  title={Automated long-term recording and analysis of neural activity in behaving animals},
  author={Dhawale, Ashesh K and Poddar, Rajesh and Wolff, Steffen BE and Normand, Valentin A and Kopelowitz, Evi and {\"O}lveczky, Bence P},
  journal={Elife},
  volume={6},
  pages={e27702},
  year={2017},
  publisher={eLife Sciences Publications, Ltd}
}

@article{azabou2023unified,
  title={A unified, scalable framework for neural population decoding},
  author={Azabou, Mehdi and Arora, Vinam and Ganesh, Venkataramana and Mao, Ximeng and Nachimuthu, Santosh and Mendelson, Michael and Richards, Blake and Perich, Matthew and Lajoie, Guillaume and Dyer, Eva},
  journal={Advances in Neural Information Processing Systems},
  volume={36},
  pages={44937--44956},
  year={2023}
}

@inproceedings{azabou2025multisession,
  title={Multi-session, Multi-task Neural Decoding from Distinct Cell-types and Brain Regions},
  author={Azabou, Mehdi and Pan, Krystal Xuejing and Arora, Vinam and Knight, Ian Jarratt and Dyer, Eva L. and Richards, Blake Aaron},
  booktitle={The Thirteenth International Conference on Learning Representations},
  year={2025}
}

@article{ye2023neural,
  title={Neural data transformer 2: multi-context pretraining for neural spiking activity},
  author={Ye, Joel and Collinger, Jennifer and Wehbe, Leila and Gaunt, Robert},
  journal={Advances in Neural Information Processing Systems},
  volume={36},
  pages={80352--80374},
  year={2023}
}

@inproceedings{karpowicz2024falcon,
  title={Few-shot Algorithms for Consistent Neural Decoding ({FALCON}) Benchmark},
  author={Karpowicz, Brianna and Ye, Joel and Fan, Chaofei and Tostado-Marcos, Pablo and Rizzoglio, Fabio and Washington, Clayton and Scodeler, Thiago and de Lucena, Diogo and Nason-Tomaszewski, Samuel and Mender, Matthew and Ma, Xuan and Arneodo, Ezequiel and Hochberg, Leigh and Chestek, Cynthia and Henderson, Jaimie and Gentner, Timothy and Gilja, Vikash and Miller, Lee and Rouse, Adam and Gaunt, Robert and Collinger, Jennifer and Pandarinath, Chethan},
  booktitle={Advances in Neural Information Processing Systems},
  volume={37},
  year={2024}
}

@article{schneider2023learnable,
  title={Learnable latent embeddings for joint behavioural and neural analysis},
  author={Schneider, Steffen and Lee, Jin Hwa and Mathis, Mackenzie Weygandt},
  journal={Nature},
  volume={617},
  number={7960},
  pages={360--368},
  year={2023},
  publisher={Nature Publishing Group UK London}
}

@article{zhang2025neural,
  title={Neural Encoding and Decoding at Scale},
  author={Zhang, Yizi and Wang, Yanchen and Azabou, Mehdi and Andre, Alexandre and Wang, Zixuan and Lyu, Hanrui and Laboratory, The International Brain and Dyer, Eva and Paninski, Liam and Hurwitz, Cole},
  journal={arXiv preprint arXiv:2504.08201},
  year={2025}
}

@article{ibl2025brainwide,
  title={A brain-wide map of neural activity during complex behaviour},
  author={Angelaki, Dora and Benson, Brandon and Benson, Julius and Birman, Daniel and Bonacchi, Niccol{\`o} and Bougrova, Kc{\'e}nia and Bruijns, Sebastian A and Carandini, Matteo and Catarino, Joana A and others},
  journal={Nature},
  volume={645},
  number={8079},
  pages={177--191},
  year={2025},
  publisher={Nature Publishing Group UK London}
}

@article{steinmetz2019distributed,
  title={Distributed coding of choice, action and engagement across the mouse brain},
  author={Steinmetz, Nicholas A and Zatka-Haas, Peter and Carandini, Matteo and Harris, Kenneth D},
  journal={Nature},
  volume={576},
  number={7786},
  pages={266--273},
  year={2019},
  publisher={Nature Publishing Group UK London}
}

@article{edelman1998geometry,
  title={The geometry of algorithms with orthogonality constraints},
  author={Edelman, Alan and Arias, Tom{\'a}s A and Smith, Steven T},
  journal={SIAM journal on Matrix Analysis and Applications},
  volume={20},
  number={2},
  pages={303--353},
  year={1998},
  publisher={SIAM}
}

@inproceedings{sennrich2016neural,
  title={Neural machine translation of rare words with subword units},
  author={Sennrich, Rico and Haddow, Barry and Birch, Alexandra},
  booktitle={Proceedings of the 54th annual meeting of the association for computational linguistics (volume 1: long papers)},
  pages={1715--1725},
  year={2016}
}

@inproceedings{kudo2018sentencepiece,
  title={SentencePiece: A simple and language independent subword tokenizer and detokenizer for neural text processing},
  author={Kudo, Taku and Richardson, John},
  booktitle={Proceedings of the 2018 conference on empirical methods in natural language processing: System demonstrations},
  pages={66--71},
  year={2018}
}

@article{churchland2012neural,
  title={Neural population dynamics during reaching},
  author={Churchland, Mark M and Cunningham, John P and Kaufman, Matthew T and Foster, Justin D and Nuyujukian, Paul and Ryu, Stephen I and Shenoy, Krishna V},
  journal={Nature},
  volume={487},
  number={7405},
  pages={51--56},
  year={2012},
  publisher={Nature Publishing Group UK London}
}

@article{gallego2017neural,
  title={Neural manifolds for the control of movement},
  author={Gallego, Juan A and Perich, Matthew G and Miller, Lee E and Solla, Sara A},
  journal={Neuron},
  volume={94},
  number={5},
  pages={978--984},
  year={2017},
  publisher={Elsevier}
}

@article{vyas2020computation,
  title={Computation through neural population dynamics},
  author={Vyas, Saurabh and Golub, Matthew D and Sussillo, David and Shenoy, Krishna V},
  journal={Annual review of neuroscience},
  volume={43},
  number={1},
  pages={249--275},
  year={2020},
  publisher={Annual Reviews}
}

@article{cunningham2014dimensionality,
  title={Dimensionality reduction for large-scale neural recordings},
  author={Cunningham, John P and Yu, Byron M},
  journal={Nature neuroscience},
  volume={17},
  number={11},
  pages={1500--1509},
  year={2014},
  publisher={Nature Publishing Group US New York}
}

@inproceedings{vaswani2017attention,
  title={Attention Is All You Need},
  author={Vaswani, Ashish and Shazeer, Noam and Parmar, Niki and Uszkoreit, Jakob and Jones, Llion and Gomez, Aidan N. and Kaiser, {\L}ukasz and Polosukhin, Illia},
  booktitle={Advances in Neural Information Processing Systems},
  volume={30},
  pages={5998--6008},
  year={2017}
}

@inproceedings{gu2024mamba,
title={Mamba: Linear-Time Sequence Modeling with Selective State Spaces},
author={Albert Gu and Tri Dao},
booktitle={First Conference on Language Modeling},
year={2024},
url={https://openreview.net/forum?id=tEYskw1VY2}
}

@article{ibl2022reproducibility,
  title={Reproducibility of In-vivo Electrophysiological Measurements in Mice},
  author={{The International Brain Laboratory} and Aguillon-Rodriguez, Valeria and Angelaki, Dora and Bayer, Hannah and Bonacchi, Niccol{\`o} and Carandini, Matteo and Cazettes, Fanny and Chapuis, Gaelle and Churchland, Anne K. and Dan, Yang and Dewitt, Eric and Faulkner, Mayo and Forber, Hamish and Haetzel, Laura and Hausser, Michael and Hofer, Sonja B. and Hu, Fei and Khanal, Anup and Krasniak, Christopher and Laranjeira, Ines and Mainen, Zachary F. and Meijer, Guido and Miska, Nathaniel J. and Mrsic-Flogel, Thomas D. and Murakami, Masayoshi and Noel, Jean-Paul and Pan-Vazquez, Alejandro and Rossant, Cyrille and Sanders, Joshua and Socha, Karolina and Terry, Rebecca and Urai, Anne E. and Vergara, Hernando and Wells, Miles and Wilson, Christian J. and Witten, Ilana B. and Wool, Lauren E. and Zador, Anthony M.},
  journal={bioRxiv},
  year={2022},
  doi={10.1101/2022.05.09.491042},
  note={doi: 10.1101/2022.05.09.491042}
}

@article{steinmetz2021neuropixels,
  title={Neuropixels 2.0: A miniaturized high-density probe for stable, long-term brain recordings},
  author={Steinmetz, Nicholas A and Aydin, Cagatay and Lebedeva, Anna and Okun, Michael and Pachitariu, Marius and Bauza, Marius and Beau, Maxime and Bhagat, Jai and B{\"o}hm, Claudia and Broux, Martijn and others},
  journal={Science},
  volume={372},
  number={6539},
  pages={eabf4588},
  year={2021},
  publisher={American Association for the Advancement of Science}
}

@article{stringer2019spontaneous,
  title={Spontaneous behaviors drive multidimensional, brainwide activity},
  author={Stringer, Carsen and Pachitariu, Marius and Steinmetz, Nicholas and Reddy, Charu Bai and Carandini, Matteo and Harris, Kenneth D},
  journal={Science},
  volume={364},
  number={6437},
  pages={eaav7893},
  year={2019},
  publisher={American Association for the Advancement of Science}
}

@article{musall2019single,
  title={Single-trial neural dynamics are dominated by richly varied movements},
  author={Musall, Simon and Kaufman, Matthew T and Juavinett, Ashley L and Gluf, Steven and Churchland, Anne K},
  journal={Nature neuroscience},
  volume={22},
  number={10},
  pages={1677--1686},
  year={2019},
  publisher={Nature Publishing Group US New York}
}

@inproceedings{pei2021neural,
  title={Neural Latents Benchmark '21: Evaluating latent variable models of neural population activity},
  author={Felix Pei and Joel Ye and David M. Zoltowski and Anqi Wu and Raeed H. Chowdhury and Hansem Sohn and Joseph E. O’Doherty and Krishna V. Shenoy and Matthew T. Kaufman and Mark Churchland and Mehrdad Jazayeri and Lee E. Miller and Jonathan Pillow and Il Memming Park and Eva L. Dyer and Chethan Pandarinath},
  booktitle={Advances in Neural Information Processing Systems (NeurIPS), Track on Datasets and Benchmarks},
  year={2021},
  url={https://arxiv.org/abs/2109.04463}
}

@article{jun2017fully,
  title={Fully integrated silicon probes for high-density recording of neural activity},
  author={Jun, James J and Steinmetz, Nicholas A and Siegle, Joshua H and Denman, Daniel J and Bauza, Marius and Barbarits, Brian and Lee, Albert K and Anastassiou, Costas A and Andrei, Alexandru and Ayd{\i}n, {\c{C}}a{\u{g}}atay and others},
  journal={Nature},
  volume={551},
  number={7679},
  pages={232--236},
  year={2017},
  publisher={Nature Publishing Group UK London}
}

@article{gallego2020long,
  title={Long-term stability of cortical population dynamics underlying consistent behavior},
  author={Gallego, Juan A and Perich, Matthew G and Chowdhury, Raeed H and Solla, Sara A and Miller, Lee E},
  journal={Nature neuroscience},
  volume={23},
  number={2},
  pages={260--270},
  year={2020},
  publisher={Nature Publishing Group US New York}
}

@article{safaie2023preserved,
  title={Preserved neural dynamics across animals performing similar behaviour},
  author={Safaie, Mostafa and Chang, Joanna C and Park, Junchol and Miller, Lee E and Dudman, Joshua T and Perich, Matthew G and Gallego, Juan A},
  journal={Nature},
  volume={623},
  number={7988},
  pages={765--771},
  year={2023},
  publisher={Nature Publishing Group UK London}
}

@article{elsayed2016reorganization,
  title={Reorganization between preparatory and movement population responses in motor cortex},
  author={Elsayed, Gamaleldin F and Lara, Antonio H and Kaufman, Matthew T and Churchland, Mark M and Cunningham, John P},
  journal={Nature communications},
  volume={7},
  number={1},
  pages={13239},
  year={2016},
  publisher={Nature Publishing Group UK London}
}

@article{dao2024transformers,
  title={Transformers are SSMs: Generalized models and efficient algorithms through structured state space duality},
  author={Dao, Tri and Gu, Albert},
  journal={arXiv preprint arXiv:2405.21060},
  year={2024}
}

@article{zacks2007event,
  title={Event perception: A mind--brain perspective},
  author={Zacks, Jeffrey M and Speer, Nicole K and Swallow, Khena M and Braver, Todd S and Reynolds, Jeremy R},
  journal={Psychological Bulletin},
  volume={133},
  number={2},
  pages={273--293},
  year={2007},
  publisher={American Psychological Association}
}

@inproceedings{jaegle2021perceiver,
  title={Perceiver: General Perception with Iterative Attention},
  author={Jaegle, Andrew and Gimeno, Felix and Brock, Andrew and Vinyals, Oriol and Zisserman, Andrew and Carreira, Joao},
  booktitle={Proceedings of the 38th International Conference on Machine Learning},
  series={Proceedings of Machine Learning Research},
  volume={139},
  pages={4651--4664},
  year={2021},
  publisher={PMLR}
}

@article{pachitariu2024suite2p,
  title={Solving the spike sorting problem with {Kilosort}},
  author={Pachitariu, Marius and Sridhar, Shashwat and Pennington, Jake and Stringer, Carsen},
  journal={Nature Methods},
  year={2024},
  publisher={Nature Publishing Group}
}

@article{linderman2017bayesian,
  title={Bayesian learning and inference in recurrent switching linear dynamical systems},
  author={Linderman, Scott and Johnson, Matthew and Miller, Andrew and Adams, Ryan and Blei, David and Paninski, Liam},
  journal={Proceedings of the 20th International Conference on Artificial Intelligence and Statistics},
  year={2017}
}

@inproceedings{yu2020learning,
  title     = {Learning Diverse and Discriminative Representations via the Principle of Maximal Coding Rate Reduction},
  author    = {Yu, Yaodong and Chan, Kwan Ho Ryan and You, Chong and Song, Chaobing and Ma, Yi},
  booktitle = {Advances in Neural Information Processing Systems},
  volume    = {33},
  year      = {2020}
}

@inproceedings{yu2023whitebox,
  title     = {White-Box Transformers via Sparse Rate Reduction},
  author    = {Yu, Yaodong and Buchanan, Sam and Pai, Druv and Chu, Tianzhe and Wu, Ziyang and Tong, Shengbang and Haeffele, Benjamin D. and Ma, Yi},
  booktitle = {Advances in Neural Information Processing Systems},
  volume    = {36},
  year      = {2023}
}

@article{degenhart2020stabilization,
  title={Stabilization of a brain--computer interface via the alignment of low-dimensional spaces of neural activity},
  author={Degenhart, Alan D and Bishop, William E and Oby, Emily R and Tyler-Kabara, Elizabeth C and Chase, Steven M and Batista, Aaron P and Yu, Byron M},
  journal={Nature biomedical engineering},
  volume={4},
  number={7},
  pages={672--685},
  year={2020},
  publisher={Nature Publishing Group UK London}
}

@article{karpowicz2022stabilizing,
  title   = {Stabilizing brain-computer interfaces through alignment of latent dynamics},
  author  = {Karpowicz, Brianna M. and Ali, Yahia H. and Wimalasena, Lahiru N. and Sedler, Andrew R. and Keshtkaran, Mohammad Reza and Bodkin, Kevin and Ma, Xuan and Miller, Lee E. and Pandarinath, Chethan},
  journal = {bioRxiv},
  year    = {2022}
}

@inproceedings{bae2026decoding,
  title={Decoding Dynamic Visual Experience from Calcium Imaging via Cell-Pattern-Aware Pretraining},
  author={Sangyoon Bae and Mehdi Azabou and Blake Aaron Richards and Jiook Cha},
  booktitle={The Fourteenth International Conference on Learning Representations},
  year={2026},
  url={https://openreview.net/forum?id=z9kAjjRejs}
}

@inproceedings{le2025spint,
  title={{SPINT}: Spatial Permutation-Invariant Neural Transformer for Consistent Intracortical Motor Decoding},
  author={Le, Trung and Fang, Hao and Li, Jingyuan and Nguyen, Tung and Mi, Lu and Orsborn, Amy and S{\"u}mb{\"u}l, Uygar and Shlizerman, Eli},
  booktitle={Advances in Neural Information Processing Systems},
  volume={38},
  year={2025}
}
\bibliographystyle{iclr2027_conference}

\newpage
\appendix
\makeatletter
\renewcommand{\thesection}{\ifnum\value{section}>26 A\@Alph{\numexpr\value{section}-26\relax}\else\@Alph{\value{section}}\fi}
\renewcommand{\theHsection}{app.\arabic{section}}
\makeatother

\section{Neuron Identity Is Session Identity}
\label{app:neuron_overlap}

Neuron identity is session identity (Section~\ref{sec:intro}) because of how extracellular electrophysiology works. Each session records its own set of neurons, and the labels that name them never carry over to another session. We verify this on the IBL Brain Wide Map and on the Gallego et al.\ data.

\textbf{No neuron label recurs.} A per-neuron tokenizer keys its vocabulary on neuron labels. In IBL these labels never recur. Each session is an acute recording with its own probe insertion, so most sessions also record different cells. We compare the probe tracks of the 361 pairs of sessions from the same subject, using the Allen CCF coordinates of every channel from the aligned histology. Of these, 342 pairs (94.7\%) never come within 100\,$\mu$m of each other, with a median closest approach of 1.2\,mm, and 19 pairs do. The labels do not match even for those 19 pairs. Spike sorting (Kilosort~\citep{pachitariu2024suite2p}) runs on each session separately and assigns session-local cluster indices, and no registry links clusters across sessions. Across all $\binom{243}{2} = 29{,}403$ session pairs (109 subjects, median 1 session per subject) no label is shared. Of those pairs, 361 (1.2\%) come from the same subject and 29{,}042 (98.8\%) from different subjects.

\textbf{Chronic recordings.} With chronically implanted arrays, such as the Utah arrays of the Gallego et al.\ and NLB data, some cells are recorded again on later days. The data do not track them. All 111 sessions of the Gallego et al.\ data number their units from 0 within the session and carry no field that links a unit to another session. Dedicated tracking can follow part of the cells~\citep{dhawale2017automated, steinmetz2021neuropixels}, but it is rarely available and the tracked set shrinks as the recorded population turns over~\citep{gallego2020long}. For the data these models receive, the per-neuron vocabulary is therefore session-specific. The population manifold persists through the turnover~\citep{gallego2020long}, and that is what TWS encodes.

\textbf{Neuron count.} Every one of the 243 sessions has at least 128 units that pass quality control, and we keep 128 per session (\texttt{max\_units}), so in IBL the dimensionality of the raw neural space is the same for every session while its axes are unrelated neurons. The count does vary on the Gallego et al.\ data (18 to 128 units per session, median 64, after the same cap).

\textbf{Implication.} A model that represents a trial as a vector in neuron-indexed space ($\R^{N_s}$) cannot compare two sessions. Two sessions with $N_1 \neq N_2$ live in spaces of different dimension. Even when $N_1 = N_2$, the axes correspond to unrelated neurons. Per-neuron tokenization, as in POYO and NEDS, answers this with learned neuron embeddings and per-session adaptation, which write neuron identity into the representation. TWS summarizes each state over an unordered set of neurons instead, which removes neuron identity by construction. Session identity is a different matter. TWS reduces it and does not remove it: trained TWS embeddings still carry $0.49$ of the session entropy (Figure~\ref{fig:structural_invariance}A).

\section{Regime Boundaries vs.\ Event Segmentation Theory}
\label{app:est_distinction}

We use ``regime boundary'' rather than ``event boundary'' to distinguish TWS's boundaries from those in Event Segmentation Theory~\citep{zacks2007event}. EST boundaries are observer-internal, driven by prediction error, and hierarchically organized; TWS boundaries are task-defined transitions (stimulus onset set by the task, movement onset measured from behavior) that demarcate distinct computational regimes. The two concepts overlap (both mark qualitative changes in processing) but differ in origin: exogenous task structure versus endogenous perceptual inference.

\section{Extended Related Work}
\label{app:extended_related}

\textbf{Subword tokenization as an analogy.} In NLP, subword methods such as BPE~\citep{sennrich2016neural} and SentencePiece~\citep{kudo2018sentencepiece} solve the open-vocabulary problem by learning reusable units below the level of whole words, from how often character sequences occur. TWS draws an analogy. Regime boundaries define units between a single time bin and a whole trial. Every such unit marks the same stage of the task in every session, whichever neurons were recorded. BPE merges character pairs by corpus frequency; TWS segments at task events, stimulus onset and movement onset. The boundaries of a TWS token therefore carry domain knowledge rather than a statistical regularity. The Grassmannian projection that follows is the neural analogue of a shared embedding table: it maps variable-length and variable-neuron states into a fixed geometry that carries no neuron identity and less session identity than per-neuron representations.

\textbf{Behavioral meaning of the manifold.} Population dynamics change regime at task events such as movement onset~\citep{vyas2020computation, churchland2012neural} and move between distinct subspaces such as the orthogonal preparatory and movement subspaces of motor cortex~\citep{elsayed2016reorganization}. The manifold follows the task and not the animal. One monkey performing two related tasks produces less similar latent dynamics than two monkeys performing the same task~\citep{safaie2023preserved}. The manifold carries this meaning before the animal moves: while a monkey only plans a reach, its dynamics already predict which target it will reach for~\citep{safaie2023preserved}.

\textbf{Grassmannian and subspace methods.} Grassmannian manifolds (spaces of linear subspaces) have a long history in computer vision for subspace tracking and video-based recognition, where each video clip is represented as a point on $\Grass(k, d)$ and geodesic distances provide an appearance-invariant similarity measure. In neuroscience, subspace overlap metrics (principal angles, canonical correlations) are standard tools for quantifying representational similarity across conditions or sessions~\citep{elsayed2016reorganization}. TWS uses the Grassmannian not as an analysis tool but as an architectural bottleneck: projecting each state onto $\Grass(k_\mathrm{dim}, d)$ with $k_\mathrm{dim}{=}2$ forces the representation through a two-dimensional subspace per state. The unordered aggregation removes neuron identity and the narrow bottleneck limits how much session-specific detail passes through. The Grassmannian is one lightweight way to impose this bottleneck.

\textbf{Neural state segmentation.} Hidden Markov models (HMMs) and switching linear dynamical systems have been used to discover latent neural states from continuous recordings~\citep{linderman2017bayesian}. These methods learn state boundaries from data. That serves exploratory analysis, but it does not serve cross-session transfer: boundaries learned on one neuron set need not align with those learned on another. TWS takes task-defined regime boundaries instead, stimulus onset and movement onset, which mark the same stage of the task in every session. The price is that TWS needs trial-structured data with known event times.

Appendix~\ref{app:est_distinction} contrasts TWS with Event Segmentation Theory.

\section{Alignment-Based Baselines versus Tokenization}
\label{app:alignment_vs_tokenization}

Contrastive embeddings such as CEBRA~\citep{schneider2023learnable} are trained on each session and compared afterwards. Manifold-alignment methods~\citep{gallego2020long, degenhart2020stabilization, karpowicz2022stabilizing} map a new session's latent trajectories onto a reference with a transform fitted to that session's data. Both buy invariance by fitting something to every new session. TWS has it by construction. Each token is a basis vector of a subspace that attention over the unordered neurons computes from the state's population activity, and no permutation of the neurons changes it. The task-defined boundaries then make the $k$-th token stand for the same computational regime in every session. A new session needs no fitting and no alignment data. The same forward pass encodes it, even at random initialization (Table~\ref{tab:main}). The price is that TWS needs the event times that define the boundaries, which alignment methods do not.

The baselines in Table~\ref{tab:main} differ in what they fit to each session. CEBRA trains one supervised and one self-supervised encoder per session, 486 fits of 5{,}000 iterations for 243 sessions. POYO learns an embedding for every neuron and every session during pretraining, over all IBL sessions, so its test-session embeddings were fitted to those sessions' data. TWS fits nothing per session. Under the same leave-one-session-out probe it decodes cross-session movement at MCC $0.565$ and RT at $R^2{=}0.402$, and at random initialization still at $0.274$ and $0.083$, while CEBRA and POYO stay at chance on both targets (Table~\ref{tab:main}). Per-session fitting makes the embedding space of one session internally consistent and leaves the axes of two sessions unrelated, which is what a shared linear probe cannot use. TWS makes the axes comparable before any probe runs. Finetuning POYO end-to-end on IBL choice and RT does not change this. Its embeddings still decode cross-session movement at the majority rate (accuracy $0.748$, MCC $0.018$; RT $R^2{=}0.036$; one model, scored on all 243 sessions).

\textbf{A label-free alignment baseline.} We implement alignment in the one form that runs on an unseen session. Per session, the log-transformed, z-scored spike counts are reduced to 10 principal components. Each session's trial-averaged latent trajectory, over all trials and aligned to the stimulus, is rotated onto a common template by orthogonal Procrustes, and generalized Procrustes over all sessions builds that template. Single trials are then projected, rotated and averaged in ten 200\,ms windows. The procedure uses no labels, and the held-out session enters through its own unlabeled trials alone. Under the probe of Table~\ref{tab:main}, aligned PCA decodes movement at MCC $0.349$ and RT at $R^2{=}0.226$, and choice and block at chance. Without the rotation the same features decode neither target ($0.002$ and $0.003$), so the rotation is what makes PCA transfer. TWS exceeds aligned PCA by $0.22$ in movement MCC and $0.18$ in RT $R^2$ and fits nothing to the new session (Table~\ref{tab:session_stats}).

\section{Empirical Validation of the Type A / Type B Distinction}
\label{app:type_ab_variability}

A variable is Type~A when it survives a change of the recorded neurons. We measure that inside a session, before
any model enters. The neurons of a session go into two random halves. Every trial of a half is summarized by
statistics that do not depend on which neuron is which: the mean, the standard deviation and the 10, 25, 50, 75
and 90\% quantiles over neurons, in five windows of the trial. A decoder is fit on the summaries of one half and
scored on the summaries of the other, with the trials cross-validated, so training and test differ in both trials
and neurons. The index divides this cross-half score by the score of a decoder that sees the full neuron vector of
the same trials. It therefore reports the share of the decodable signal that survives the loss of neuron identity,
not the strength of that signal. Every neuron is z-scored within its session first, as in TWS, and the summaries
of each half are standardized over the trials of that half, without labels. Ten neuron splits and five folds per
session, over 242 IBL sessions, 39 Steinmetz sessions, 81 Gallego et al.\ sessions and 23 NLB pseudo-sessions.
The index needs 60 trials for its folds and 10 trials in the smallest class,
which drops one IBL session of 50 trials. Reach direction exists only in the 84 center-out sessions of Gallego et
al., since the random-target sessions draw a new target on every trial; three of those 84 fall below the trial or
neuron minimum.

\textbf{Results} (Table~\ref{tab:type_ab_index}). The index separates the movement targets from IBL choice and block. Over the eight pairs it follows the score of TWS from scratch only loosely (Spearman $0.55$, $p{=}0.16$). Movement keeps more of its signal than choice or block on IBL and Steinmetz, and NLB movement keeps about a third of it. Choice on IBL keeps a quarter of it. Steinmetz choice keeps about 1.7 times as much as IBL choice, which is why the same target name transfers on one
dataset and not on the other. The Steinmetz label contrasts a left turn with a right turn or no turn
(Table~\ref{tab:boundary_targets}), so part of it follows population-wide movement. The cross-half column and
Table~\ref{tab:cross_dataset} share a metric: on Gallego et al.\ the identity-free decoder achieves $0.112$
within a session, and TWS achieves $0.232$ across sessions, species and recording technologies. Reaction time has
no index, because the full-neuron decoder gives a negative $R^2$ within sessions and leaves the ratio undefined.

\begin{table}[!ht]
\centering
\caption{How much of each target survives a change of the recorded neurons. MCC, averaged over sessions.}
\label{tab:type_ab_index}
\small
\begin{tabular}{llccc}
\toprule
Dataset & Target & Full neurons & Cross-half summary & Index \\
\midrule
\multirow{3}{*}{IBL} & movement & $0.524$ & $0.308$ & $0.59$ \\
 & choice & $0.624$ & $0.155$ & $0.25$ \\
 & block & $0.560$ & $0.143$ & $0.25$ \\
\midrule
\multirow{3}{*}{Steinmetz} & movement & $0.804$ & $0.611$ & $\mathbf{0.76}$ \\
 & feedback & $0.477$ & $0.287$ & $0.60$ \\
 & choice & $0.807$ & $0.345$ & $0.43$ \\
\midrule
NLB & movement & $0.663$ & $0.234$ & $0.35$ \\
Gallego et al. & reach direction & $0.952$ & $0.112$ & $\mathbf{0.12}$ \\
\bottomrule
\end{tabular}

\end{table}

\section{Training Details and Hyperparameters}
\label{app:training}

\textbf{Geometry-only pretraining.} Pretraining sets up the geometry of the token space before behavior is learned. Preparatory and movement-epoch activity occupy orthogonal subspaces~\citep{elsayed2016reorganization}, so we push the subspaces of consecutive states apart and let behavioral training fill a space that already has this structure; Section~\ref{sec:tradeoff} shows that it adds session invariance rather than accuracy. We train with three self-supervised losses, no behavioral labels:
\begin{equation}
    \Ls = \underbrace{\Ls_\mathrm{MSE}}_{\text{reconstruct epoch means}} + \lambda_\mathrm{div} \underbrace{\Ls_\mathrm{div}}_{\text{state token diversity}} + \lambda_\mathrm{geo} \underbrace{\Ls_\mathrm{geo}}_{\text{geodesic distance}}
    \label{eq:pretrain_loss}
\end{equation}

$\Ls_\mathrm{MSE}$: Mean squared error between decoded state tokens and per-state neuron mean activities. $\Ls_\mathrm{div}$: Chordal distance (squared sine of principal angles between consecutive subspaces) plus pairwise cosine repulsion across all state token pairs, jointly encouraging manifold separation. $\Ls_\mathrm{geo}$: Negative geodesic distance on the Grassmannian between adjacent state pairs ($S_k, S_{k+1}$), encouraging near-orthogonality between temporally neighboring regimes, the prior stated above.

\textbf{State masking pretext.} During pretraining, we randomly mask one of $K$ states (replacing its tokens with a learnable mask embedding) in 50\% of trials. The model must reconstruct the masked state from the remaining states, learning inter-state relationships.

\textbf{Behavioral finetuning.} After pretraining, we add one linear head per token for each classification task, with the logits averaged over tokens, and regression heads on the mean token $\bar{\vh} = \mathrm{mean}(\vh) \in \R^d$, and finetune on behavioral labels. Models are evaluated both within-session (same session, held-out trials) and cross-session (test sessions never seen during training).

Table~\ref{tab:hyperparams} lists the settings of every training stage. All stages use AdamW; IBL pretraining and behavioral training clip gradients at norm $1$. Each session enters with at most 128 units. In IBL we keep the units that pass the IBL quality label and take the 128 with the most spikes over the session; every IBL session has at least 128 such units. In the other datasets we keep the 128 units with the largest variance and fill the remaining slots with zeros when a session has fewer. Pretraining further keeps IBL units that pass firing-rate, presence-ratio and contamination thresholds, cuts at the knee of their spike counts and keeps the $40\%$ whose binned counts are closest to Gaussian, a selection inspired by \citet{bae2026decoding}. No unit selection uses behavioral labels.

\textbf{Learned offsets.} The offsets of Eq.~\ref{eq:soft_weights} are trained with the other weights and bounded by a sigmoid to $\Delta_1 \in [5, 25]$ and $\Delta_2 \in [0, 20]$ bins. In the five TWS from scratch models of Table~\ref{tab:main} they converge to $\Delta_1 = 14.6$--$15.6$ bins ($292$--$311$\,ms after the stimulus onset) and $\Delta_2 = 1.3$--$3.0$ bins ($26$--$60$\,ms after the first movement). $\Delta_1$ lands after the stimulus-evoked transient and $\Delta_2$ almost on the movement itself; we read these as the delays at which each regime settles, not as a measurement of them. The architecture is the same in every stage: $d{=}256$, $K{=}3$ states with $k_\mathrm{dim}{=}2$, gate temperature $\gamma{=}2$ bins, 4 cross-attention heads, at most 128 neurons and $T{=}100$ bins of 20\,ms; the backbone has 4 blocks, each a layer normalization, a kernel-3 convolution over the 6 tokens, GELU and dropout $0.1$, followed by a feed-forward layer.

\begin{table}[!ht]
\centering
\caption{Training hyperparameters. The six IBL behavioral heads are choice, feedback, licking, movement, wheel velocity and RT. New-dataset training follows Appendix~\ref{app:xds_finetune}.}
\label{tab:hyperparams}
\small
\setlength{\tabcolsep}{4pt}
\resizebox{\textwidth}{!}{
\begin{tabular}{llll}
\toprule
 & IBL pretraining & IBL behavioral training & New-dataset training \\
\midrule
Objective & Eq.~\ref{eq:pretrain_loss}, no labels & six behavioral heads & class-weighted cross-entropy \\
Loss weights & $\lambda_\mathrm{div}{=}0.01$, $\lambda_\mathrm{geo}{=}0.005$ & -- & -- \\
State masking & $50\%$ of trials & -- & -- \\
Learning rate & $8 \times 10^{-4}$, constant & $10^{-3}$, cosine to $10^{-5}$ & encoder $10^{-4}$, head $10^{-3}$ \\
Weight decay & $0.01$ & $0.05$ & $0.01$ \\
Batch & 256 trials from 16 sessions & 128 trials from 8 sessions & 4 sessions $\times$ 16 trials \\
Epochs & 30 (patience 15) & 40 (no early stopping) & at most 15 (patience 5) \\
Validation & $5\%$ of trials & $15\%$ of sessions & $15\%$ of training sessions \\
\bottomrule
\end{tabular}
}

\end{table}

\section{Additional Ablations}
\label{app:ablations}

\textit{Backbone note.} The $k_\mathrm{dim}$ table and the first per-state angles below come from an earlier backbone on the same tokenizer (a multi-scale state-space model) and report accuracy; the other results in this section come from the held-out models of Table~\ref{tab:main}.

\subsection{Manifold Dimensionality (\texorpdfstring{$k_\mathrm{dim}$}{k\_dim})}

\begin{table}[!ht]
\caption{Manifold dimensionality ablation (IBL).}
\begin{center}
\small
\begin{tabular}{ccc}
\toprule
$k_\mathrm{dim}$ & Movement & Interpretation \\
\midrule
1 & 0.864 & Slightly narrow: marginal loss of task-relevant variance \\
\textbf{2} & \textbf{0.865} & Selected: balances expressiveness and compression \\
4 & 0.867 & Similar: extra dimensions add little at this scale \\
\bottomrule
\end{tabular}
\end{center}

\end{table}

With 243 sessions of decoder training data, movement accuracy holds across $k_\mathrm{dim}$ ($0.864$--$0.867$). All three values decode within $0.003$ of one another, and we use $k_\mathrm{dim}{=}2$.

\subsection{\texorpdfstring{Tokens without Gram--Schmidt}{Tokens without Gram-Schmidt}}
\label{app:no_gs}

We remove the Gram--Schmidt step of Algorithm~\ref{alg:tws} (Step~3). Each state then keeps the two LayerNorm outputs of its cross-attention as its tokens, and the tokens no longer form an orthonormal basis on $\Grass(k_\mathrm{dim}, d)$. Everything else stays as in TWS from scratch. Cross-session decoding changes by less than $0.01$ on every target (Table~\ref{tab:no_gs}). The Grassmannian form is therefore a convenient realization of the tokens and not a requirement.

\begin{table}[!ht]
\centering
\caption{Cross-session decoding on the 53 held-out IBL sessions (protocol of Table~\ref{tab:main}); mean $\pm$ SD over 5 model seeds.}
\label{tab:no_gs}
\small
\resizebox{\textwidth}{!}{
\begin{tabular}{lcccc}
\toprule
Tokens & Movement MCC & RT $R^2$ & Choice MCC & Block MCC \\
\midrule
Orthonormal basis (TWS from scratch) & $0.565 \pm 0.006$ & $0.402 \pm 0.003$ & $0.038 \pm 0.006$ & $0.026 \pm 0.004$ \\
Without Gram--Schmidt & $0.562 \pm 0.003$ & $0.399 \pm 0.003$ & $0.033 \pm 0.005$ & $0.022 \pm 0.005$ \\
\bottomrule
\end{tabular}
}

\end{table}

\subsection{Per-state Manifold Orthogonality}

In an earlier geometry-only pretraining run the first principal angle between consecutive state subspaces was $88.8^\circ$ for S0$\leftrightarrow$S1 and $88.8^\circ$ for S1$\leftrightarrow$S2, against $1.5^\circ$ and $5.4^\circ$ without pretraining. The pretrained subspaces are nearly orthogonal; the others overlap heavily. The pretrained model analyzed in Appendix~\ref{app:state_manifold} (with state masking, before finetuning) gives $87.0^\circ$ and $87.3^\circ$ for the first principal angle of the two consecutive pairs, over the trials of all 243 sessions. Geometric pretraining separates the state manifolds. 

\section{Backbone Architecture}
\label{app:backbone}

The backbone receives the six state tokens and nothing else. TWS has already averaged activity within each state (Algorithm~\ref{alg:tws}, Step~2), so the backbone mixes a short sequence of state summaries rather than time bins. Any sequence model can fill that role (Section~\ref{sec:backbone_recovery}). We use a convolutional mixer by default: four blocks of a kernel-3 convolution over the six state tokens, with $d{=}256$. It is the simplest backbone that mixes the tokens at all, and it leads the Mamba and Transformer backbones by at most $0.03$ in movement MCC (Table~\ref{tab:backbone_ablation}). Backbones that mix the tokens decode within $0.03$ MCC of each other. They differ in how well they transfer to other datasets (Appendix~\ref{app:backbone_sweep}).

\section{Implementation Details}
\label{app:implementation}

\textbf{Permutation invariance: proof and check.} Let $\Pi$ permute the $N_s$ recorded neurons and apply it to the activity $\mX$ and to the coordinates $\{\vp_n\}$. Then every TWS token is unchanged. \emph{Proof.} Normalization, $f_\mathrm{act}$ and $f_\mathrm{coord}$ act on each neuron alone and so commute with $\Pi$. The cross-attention returns a softmax-weighted sum over the neurons with no positional encoding over them and so ignores their order. Gram-Schmidt then acts on the $k_\mathrm{dim}$ attention outputs in a fixed order. $\square$ The backbone sees the six tokens in a fixed order so predictions and distances between tokens inherit the invariance. On five held-out sessions and 20 random permutations each the tokens of a trained model agree to within $4 \times 10^{-6}$ of their scale.

\textbf{Scope of the permutation invariance.} Each token summarizes the mean activity of every neuron within the state. The invariance holds for the order of the neurons. Recording different neurons changes the activity that enters the aggregation. The aggregation has no padding mask, so in a session with fewer than 128 units the zero-filled slots enter the attention and the tokens depend on the number of recorded units; every IBL session has 128. The backbone receives the orthonormal basis that Gram-Schmidt returns, a point on the Stiefel manifold whose span is the Grassmannian point. The basis therefore enters the representation alongside its span.

\textbf{Atlas coordinates.} In the neuron feature computation $\vu_n = f_\mathrm{coord}(\vp_n) + f_\mathrm{act}(\bar{x}_{k,n})$, atlas coordinates ($\vp_n \in \R^3$) are missing for 230 of the 243 IBL sessions, so $f_\mathrm{coord}$ reduces to one learned bias shared by all neurons. In the other 230 sessions every IBL result therefore comes from $f_\mathrm{act}$ alone: the tokens line up across sessions from per-state mean firing rates, with no anatomical identity. When atlas coordinates are available, $f_\mathrm{coord}$ provides an additional anatomical prior.

\section{Two Controls on the Head and on the Neurons}
\label{app:global_pool}

\textbf{One head over all tokens.} Behavioral training fits one linear head per state (Algorithm~\ref{alg:tws}).
Pooling the six tokens under a single head leaves cross-session decoding unchanged: movement
$0.562 \pm 0.004$ against $0.565 \pm 0.006$, reaction time $0.402 \pm 0.001$ against $0.402 \pm 0.003$, over the
53 held-out sessions and 5 model seeds. What the tokens hold does not depend on how the head aggregates them.

\textbf{Fewer neurons per session.} Every session enters with at most 128 units (Section~\ref{sec:setup}). We keep
a random subset of them, zero the rest and leave the model of Table~\ref{tab:main} unchanged. Decoding then falls
gradually rather than abruptly (Table~\ref{tab:neuron_subset}): 32 units keep $83\%$ of the movement decoding of
128 units, and 64 units keep $93\%$.

\begin{table}[!ht]
\centering
\caption{TWS from scratch on the 53 held-out sessions with fewer units per session, the rest zero-filled. Mean $\pm$ SD over 5 model seeds.}
\label{tab:neuron_subset}
\small
\begin{tabular}{lcccc}
\toprule
Units per session & Movement MCC & RT $R^2$ & Choice MCC & Block MCC \\
\midrule
32 & $0.469 \pm 0.011$ & $0.346 \pm 0.015$ & $0.018 \pm 0.006$ & $0.031 \pm 0.004$ \\
64 & $0.523 \pm 0.005$ & $0.380 \pm 0.004$ & $0.025 \pm 0.005$ & $0.022 \pm 0.005$ \\
128 (Table~\ref{tab:main}) & $\mathbf{0.565 \pm 0.006}$ & $\mathbf{0.402 \pm 0.003}$ & $\mathbf{0.038 \pm 0.006}$ & $0.026 \pm 0.004$ \\
\bottomrule
\end{tabular}

\end{table}

\section{Glossary of Behavioral Variables}
\label{app:glossary}

Table~\ref{tab:glossary} defines the seven behavioral variables used in this study. All variables are derived from the IBL biased-blocks visual decision-making task~\citep{ibl2022reproducibility, ibl2025brainwide}, in which head-fixed mice turn a steering wheel to indicate the perceived location of a drifting visual grating.

\begin{table}[!ht]
\caption{Behavioral variable definitions. Type~A: encoded across the population, recoverable from any sufficiently large neuronal subset. Type~B: depends on which specific neurons are recorded. Scale: C = classification, R = regression. $^*$Feedback is structurally Type~A but its cross-session decodability is dominated by its correlation with movement.}
\label{tab:glossary}
\begin{center}
\small
\begin{tabular}{lcccp{6.8cm}}
\toprule
Variable & Scale & Type & Base rate & Definition \\
\midrule
Movement & C & A & $P{=}0.76$ & Whether a first wheel movement (IBL \texttt{firstMovement\_times}) occurs between stimulus onset and $2$\,s after it. This is the event that also places the first-movement regime boundary, so the boundary alone largely determines the label (Appendix~\ref{app:gate_decomposition}). Movement-related neural signals appear ``almost everywhere in the brain''~\citep{musall2019single, stringer2019spontaneous}. \\[4pt]
Reaction time & R & A & --- & Time (s) from stimulus onset to the IBL response time, which follows the first movement by a median of $0.19$\,s. Trials outside $50$\,ms--$10$\,s are excluded. Reflects sensorimotor transformation latency; shorter for higher-contrast stimuli. \\[4pt]
Choice & C & B & $P{\approx}0.52$ & Direction of wheel turn: leftward (CW) or rightward (CCW). Correct when the grating is moved to center ($0^\circ$ azimuth). Requires lateralized selectivity in specific neurons~\citep{steinmetz2019distributed}. \\[4pt]
Feedback & C & A$^*$ & $P{=}0.81$ & Trial outcome: correct ($+1$, water reward) or incorrect ($-1$, white-noise burst $+$ 2\,s timeout). Mechanistically linked to movement: $P(\text{fb}{=}1 \mid \text{move}{=}1) {=} 0.845$. \\[4pt]
Block & C & B & $P{\approx}0.52$ & Dominant stimulus side in the current probability block. After 90 unbiased trials, prior probability alternates between 80:20 and 20:80 (left:right) in blocks of 20--100 trials. A latent context variable. \\[4pt]
Engagement & C & A & $P{=}0.94$ & Whether the mouse is actively performing: recent accuracy ${\geq}0.6$ over a 10-trial rolling window. Near-ceiling base rate reduces discriminative value. \\[4pt]
Wheel velocity & R & A & --- & Signed inverse latency of the first movement (1/s, sign of the choice), a measure of movement vigor. Trials without a first movement $0.05$--$2$\,s after stimulus onset are excluded. Per-session means shift widely ($\sigma_\text{cross}{=}1.62$). \\
\bottomrule
\end{tabular}
\end{center}

\end{table}

\section{Behavioral Variable Class Distributions}
\label{app:class_dist}

Five of the seven behavioral variables separate no method from another (Table~\ref{tab:main_full}). Table~\ref{tab:class_dist} gives the reason. Each of the five is at ceiling, dominated by its base rate, or unpredictable across sessions by construction.

\begin{table}[!ht]
\caption{Class distributions and base rates for the 7 IBL behavioral variables (243 sessions, 152{,}794 trials). $P(\text{positive})$: proportion of positive class. $\sigma_\text{cross}(\mu)$: std of per-session means.}
\label{tab:class_dist}
\begin{center}
\small
\resizebox{\textwidth}{!}{
\begin{tabular}{lccl}
\toprule
Variable & $P(\text{positive})$ & Majority baseline & Explanation \\
\midrule
\multicolumn{3}{l}{\textit{Classification variables}} \\
\textbf{Movement} & 0.760 & 0.760 & Session range $[0.08, 0.91]$; \textbf{discriminative} \\
Engagement & 0.938 & 0.938 & 93\% positive; MCC $\approx 0$ for all methods \\
Feedback & 0.813 & 0.813 & $P(\text{fb}{=}1 | \text{move}{=}1) {=} 0.845$; dominated by movement \\
Choice & 0.521 & 0.521 & 69\% of sessions in $[0.4, 0.6]$; Type B, $\approx$chance \\
Block & 0.518 & 0.518 & Session std $= 0.070$; Type B, $\approx$chance \\
\midrule
\multicolumn{3}{l}{\textit{Regression variables}} \\
\textbf{RT} & \multicolumn{2}{c}{$\sigma_\text{global} = 1.49$, $\sigma_\text{cross}(\mu) = 0.57$} & Meaningful cross-session variance; \textbf{discriminative} \\
Wheel vel. & \multicolumn{2}{c}{$\sigma_\text{global} = 7.63$, $\sigma_\text{cross}(\mu) = 1.62$} & large per-session shifts; $R^2 \approx 0$ across sessions \\
\bottomrule
\end{tabular}
}
\end{center}

\end{table}

\textbf{Engagement} ($P{=}0.938$): predicting ``engaged'' for every trial yields 93.8\% accuracy, and no method decodes it across sessions (MCC $\approx 0$). \textbf{Feedback} ($P{=}0.813$) follows movement ($P(\text{fb}{=}1 \mid \text{move}{=}1) {=} 0.845$), yet no method decodes it above MCC $0.11$ across sessions. \textbf{Choice and block} ($P \approx 0.52$) are Type B. The Grassmannian bottleneck drops the neuron-specific patterns that encode them, so every cross-session method lands at chance, while per-neuron methods within a session confirm the information exists in the raw data (firing rates and Ridge: choice MCC $0.537$, block $0.574$; Table~\ref{tab:main_full}). \textbf{Wheel velocity} gives $R^2 \approx 0$ across sessions for every method. Its per-session means shift widely ($\sigma_\text{cross} {=} 1.62$).

\section{Full Results Across All Behavioral Variables}
\label{app:full_results}

Table~\ref{tab:main_full} reports cross-session and within-session decoding of all 7 IBL behavioral variables (53 held-out sessions, 5 seeds), with the cross-session protocol of Table~\ref{tab:main}. Within-session: a random 60/20/20 split of the trials so that training and test trials share sessions. TWS and the other shared representations use one probe on the trials of the 53 held-out sessions together. The per-neuron references (PCA, firing rates and CEBRA) use one probe per session because their features differ between sessions. Per-variable class distributions and chance levels are detailed in Appendix~\ref{app:class_dist}.

\begin{table}[!ht]
\caption{Full results on all 7 IBL behavioral variables. MCC for classification; $R^2$ for wheel velocity and RT, with RT standardized per session across sessions and raw within sessions. Cross-session: leave-one-session-out Ridge probe, scored on the 53 held-out sessions. Within-session: TWS and PerceiverIO are probed on the trials of the 53 held-out sessions pooled (random 60/20/20 split, 5 repeats), which none of them trained on; POYO and Population Statistics on the trials of all sessions pooled; PCA, firing rates and CEBRA, whose feature spaces differ between sessions, per session (random 60/20/20 split). $\pm$: SD over 5 model seeds; rows without $\pm$ are a single run. PerceiverIO uses the same CNN backbone as TWS. Aligned and unaligned PCA are scored on four variables only (Appendix~\ref{app:alignment_vs_tokenization}). Per-variable class distributions in Appendix~\ref{app:class_dist}.}
\label{tab:main_full}
\begin{center}
\resizebox{\textwidth}{!}{
\begin{tabular}{lccccccc}
\toprule
Method & Movement & Feedback & Choice & Block & Engagement & Wheel ($R^2$) & RT ($R^2$) \\
\midrule
\multicolumn{8}{l}{\textit{Cross-session (leave-one-session-out)}} \\
TWS pretrained + finetuned & $0.553 \pm 0.006$ & $0.070 \pm 0.012$ & $\mathbf{0.040 \pm 0.006}$ & $\mathbf{0.031 \pm 0.003}$ & $0.000 \pm 0.000$ & $-0.000 \pm 0.001$ & $0.398 \pm 0.004$ \\
TWS from scratch & $\mathbf{0.565 \pm 0.006}$ & $0.065 \pm 0.011$ & $0.038 \pm 0.006$ & $0.026 \pm 0.004$ & $0.000 \pm 0.000$ & $-0.001 \pm 0.003$ & $\mathbf{0.402 \pm 0.003}$ \\
TWS pretrained & $0.428 \pm 0.043$ & $0.019 \pm 0.019$ & $0.022 \pm 0.001$ & $0.019 \pm 0.002$ & $-0.000 \pm 0.000$ & $-0.001 \pm 0.001$ & $0.247 \pm 0.036$ \\
TWS (random init) & $0.274 \pm 0.018$ & $0.001 \pm 0.002$ & $-0.000 \pm 0.004$ & $0.005 \pm 0.005$ & $0.002 \pm 0.003$ & $-0.002 \pm 0.001$ & $0.083 \pm 0.011$ \\
PerceiverIO & $0.303 \pm 0.009$ & $\mathbf{0.102 \pm 0.012}$ & $0.014 \pm 0.009$ & $0.011 \pm 0.006$ & $0.000 \pm 0.000$ & $-0.005 \pm 0.003$ & $0.169 \pm 0.003$ \\
PCA (aligned) & $0.349$ & -- & $0.000$ & $-0.005$ & -- & -- & $0.226$ \\
PCA (unaligned) & $0.002$ & -- & $0.027$ & $0.007$ & -- & -- & $0.003$ \\
POYO & $-0.000 \pm 0.001$ & $-0.000 \pm 0.000$ & $-0.000 \pm 0.002$ & $0.000 \pm 0.002$ & $0.000 \pm 0.000$ & $-0.000 \pm 0.001$ & $-0.001 \pm 0.001$ \\
CEBRA & $0.000$ & $0.000$ & $0.005$ & $-0.012$ & $0.000$ & $-0.000$ & $0.000$ \\
Population Statistics & $0.000$ & $0.000$ & $0.005$ & $-0.002$ & $0.000$ & $-0.000$ & $0.009$ \\
\textit{Event-time information} & \textit{0.908} & \textit{0.000} & \textit{0.031} & \textit{0.020} & \textit{0.000} & \textit{0.003} & \textit{0.443} \\
\midrule
\multicolumn{8}{l}{\textit{Within-session}} \\
TWS pretrained + finetuned & $0.585 \pm 0.005$ & $0.092 \pm 0.034$ & $0.106 \pm 0.005$ & $0.095 \pm 0.005$ & $0.000 \pm 0.000$ & $0.014 \pm 0.001$ & $0.543 \pm 0.003$ \\
TWS from scratch & $\mathbf{0.606 \pm 0.004}$ & $0.107 \pm 0.021$ & $0.109 \pm 0.006$ & $0.096 \pm 0.007$ & $0.000 \pm 0.000$ & $0.013 \pm 0.002$ & $\mathbf{0.547 \pm 0.004}$ \\
TWS pretrained & $0.448 \pm 0.040$ & $0.030 \pm 0.023$ & $0.081 \pm 0.006$ & $0.067 \pm 0.010$ & $-0.000 \pm 0.000$ & $0.009 \pm 0.001$ & $0.354 \pm 0.044$ \\
TWS (random init) & $0.360 \pm 0.021$ & $0.067 \pm 0.005$ & $0.050 \pm 0.004$ & $0.047 \pm 0.001$ & $\mathbf{0.031 \pm 0.012}$ & $0.005 \pm 0.001$ & $0.250 \pm 0.018$ \\
PerceiverIO & $0.353 \pm 0.008$ & $\mathbf{0.150 \pm 0.014}$ & $0.102 \pm 0.008$ & $0.069 \pm 0.008$ & $0.000 \pm 0.000$ & $\mathbf{0.018 \pm 0.002}$ & $0.254 \pm 0.006$ \\
POYO & $0.013 \pm 0.012$ & $0.001 \pm 0.002$ & $0.041 \pm 0.011$ & $0.026 \pm 0.007$ & $-0.000 \pm 0.000$ & $0.005 \pm 0.002$ & $0.013 \pm 0.003$ \\
CEBRA & $0.330$ & $0.090$ & $\mathbf{0.132}$ & $\mathbf{0.192}$ & $0.011$ & $0.017$ & $0.147$ \\
Population Statistics & $0.007$ & $0.006$ & $0.007$ & $0.004$ & $-0.001$ & $0.000$ & $0.015$ \\
\addlinespace[3pt]
\multicolumn{8}{l}{\textit{Reference: within-session, per-neuron features}} \\
PCA + Ridge & $\mathbf{0.469}$ & $0.382$ & $0.512$ & $0.440$ & $0.043$ & $\mathbf{0.344}$ & $\mathbf{0.378}$ \\
Firing Rate + Ridge & $0.370$ & $\mathbf{0.498}$ & $\mathbf{0.537}$ & $\mathbf{0.574}$ & $\mathbf{0.203}$ & $0.284$ & $0.354$ \\
\bottomrule
\end{tabular}
}
\end{center}

\end{table}

\section{Training on New Datasets: Initialization and Frozen Encoders}
\label{app:xds_finetune}

For each dataset the sessions are split into 5 folds; each fold's sessions are held out for testing, $15\%$ of the remaining sessions are used for early stopping, and the rest for training (hyperparameters in Table~\ref{tab:hyperparams}). The head is a layer normalization followed by a linear classifier on the mean of the output tokens. \emph{Scratch}: all weights randomly initialized. \emph{IBL}: initialized from the self-supervised IBL checkpoint (TWS) or from the IBL PerceiverIO trained on IBL behavior (lr $10^{-4}$). \emph{Frozen}: IBL encoder fixed, only the head trained. The POYO row of Table~\ref{tab:cross_dataset} uses this protocol: POYO trains from scratch on the labels of each dataset, with unit and session embeddings learned on the training sessions, and is scored on the held-out sessions of each fold, whose units it has never seen. The CEBRA row fits one supervised model per session on the trial-mean firing rates and scores its embeddings with the leave-one-session-out Ridge probe of the other rows. \emph{Frozen, behavior}: the same, with the encoders of Tables~\ref{tab:cross_dataset} and~\ref{tab:xds_pio} on both sides, trained on IBL behavior from scratch with the CNN backbone and without the 53 held-out sessions, so the two tokenizers are matched. MCC is computed per test session and averaged over sessions and folds; Table~\ref{tab:xds_finetune} gives mean $\pm$ SD over 3 seeds.

\begin{table}[!ht]
\centering
\caption{Held-out-session MCC after training on each dataset. In the IBL and Frozen columns the initializations are not matched: TWS starts from the self-supervised checkpoint, PerceiverIO from a model trained on IBL behavior with a Transformer backbone. Frozen, behavior: both encoders trained on IBL behavior with the CNN backbone.}
\label{tab:xds_finetune}
\small
\resizebox{\textwidth}{!}{
\begin{tabular}{llcccccccc}
\toprule
& & \multicolumn{4}{c}{TWS} & \multicolumn{4}{c}{PerceiverIO} \\
\cmidrule(lr){3-6}\cmidrule(lr){7-10}
Dataset & Target & Scratch & IBL & Frozen & \shortstack{Frozen,\\behavior} & Scratch & IBL & Frozen & \shortstack{Frozen,\\behavior} \\
\midrule
Steinmetz & Choice & $\mathbf{0.261 \pm 0.005}$ & $0.238 \pm 0.002$ & $0.221 \pm 0.004$ & $0.230 \pm 0.018$ & $0.231 \pm 0.007$ & $0.252 \pm 0.015$ & $0.232 \pm 0.019$ & $0.218 \pm 0.011$ \\
Gallego et al. & Reach direction & $0.234 \pm 0.003$ & $0.225 \pm 0.015$ & $0.166 \pm 0.023$ & $0.200 \pm 0.002$ & $0.281 \pm 0.007$ & $\mathbf{0.300 \pm 0.008}$ & $0.175 \pm 0.007$ & $0.183 \pm 0.005$ \\
NLB & Movement & $0.311 \pm 0.013$ & $0.354 \pm 0.010$ & $0.333 \pm 0.019$ & $0.333 \pm 0.011$ & $0.299 \pm 0.043$ & $\mathbf{0.421 \pm 0.013}$ & $0.379 \pm 0.008$ & $0.399 \pm 0.008$ \\
\bottomrule
\end{tabular}
}

\end{table}

\textbf{Within-state time and session identity come together.} Trained on each dataset (Table~\ref{tab:xds_finetune}), TWS leads on Steinmetz choice from scratch, and PerceiverIO leads on reach direction (Gallego et al.) and, from IBL behavior, on NLB; under the Ridge probe of Table~\ref{tab:xds_pio} the order on choice and reach is reversed. With both encoders trained on IBL behavior and frozen, TWS leads on choice and reach direction and PerceiverIO on NLB movement (Frozen, behavior). One reading follows from the design. The trajectory of population activity during the movement carries a reach. TWS averages that trajectory within the post-movement state, and PerceiverIO keeps it in time patches. Session identity lives in the same time course: session identity explains $31\%$ of the variance of trained PerceiverIO embeddings against $23\%$ for TWS, and movement, after each session's mean is removed, $1\%$ against $11\%$ (Figure~\ref{fig:representation}B). In the tokenizers we compared, a representation that kept within-state time also kept the session. TWS gives up within-state time and keeps less session identity ($23\%$ against $31\%$), not none. The IBL checkpoint adds to training from scratch on NLB ($+0.043$), the dataset farthest from IBL. It costs a little on Steinmetz ($-0.023$), the mouse task IBL already resembles, and on Gallego et al.\ ($-0.009$).

\section{CEBRA Regression Performance}
\label{app:cebra_regression}

Without per-session standardization of RT, CEBRA gives a negative cross-session $R^2$ on wheel velocity ($-0.057$) and on RT ($-0.191$). Self-supervised CEBRA gives the same picture ($-0.056$ and $-0.188$). Per-session contrastive embeddings carry no continuous variable across sessions. With RT standardized per session, as in Table~\ref{tab:main}, the saved embeddings of one set of CEBRA models give $R^2{=}{-}0.001$; the same embeddings without standardization give $-0.195$, against $-0.191$ for the 5-seed mean above.

\section{Physiological Variable Decoding}
\label{app:physio}

Five IBL sessions in our set have video of the whisker pad and the pupil. All five are among the 53 held-out sessions. We decode whisker motion energy and pupil diameter per trial from the frozen embeddings of the held-out models, with the Ridge probe of Table~\ref{tab:main}. Cross-session leaves one of the five sessions out; within-session splits the trials of the five at random. No model trains on either variable (Table~\ref{tab:physio}).

\begin{table}[!ht]
\centering
\caption{Whisker motion energy and pupil diameter ($R^2$) on the 5 held-out IBL sessions with video. Mean $\pm$ SD over 5 model seeds.}
\label{tab:physio}
\small
\resizebox{\textwidth}{!}{
\begin{tabular}{lcccc}
\toprule
 & \multicolumn{2}{c}{Cross-session} & \multicolumn{2}{c}{Within-session} \\
\cmidrule(lr){2-3}\cmidrule(lr){4-5}
Method & Whisker & Pupil & Whisker & Pupil \\
\midrule
TWS from scratch & $\mathbf{0.462 \pm 0.015}$ & $-0.033 \pm 0.008$ & $\mathbf{0.636 \pm 0.012}$ & $0.549 \pm 0.013$ \\
TWS pretrained + finetuned & $0.459 \pm 0.018$ & $-0.035 \pm 0.009$ & $0.607 \pm 0.018$ & $0.521 \pm 0.045$ \\
TWS pretrained & $0.090 \pm 0.036$ & $-0.035 \pm 0.002$ & $0.499 \pm 0.026$ & $0.329 \pm 0.097$ \\
TWS (random init) & $0.091 \pm 0.016$ & $\mathbf{0.008 \pm 0.001}$ & $0.306 \pm 0.066$ & $0.116 \pm 0.145$ \\
PerceiverIO & $0.227 \pm 0.034$ & $-0.072 \pm 0.081$ & $0.619 \pm 0.019$ & $0.751 \pm 0.012$ \\
POYO & $-0.081 \pm 0.148$ & $-0.039 \pm 0.054$ & $0.452 \pm 0.057$ & $\mathbf{0.835 \pm 0.080}$ \\
\bottomrule
\end{tabular}
}

\end{table}

Across sessions TWS decodes whisker motion twice as well as PerceiverIO, and POYO decodes none of it. Behavioral training produces the result: TWS (random init) and TWS pretrained both stay near $0.09$. No model decodes pupil diameter across sessions, and POYO and PerceiverIO decode it well within sessions. Five sessions is a small sample, so we read these numbers as a direction only.

\section{State Manifold Characteristics}
\label{app:state_manifold}

\begin{figure}[!ht]
\centering
\includegraphics[width=\textwidth]{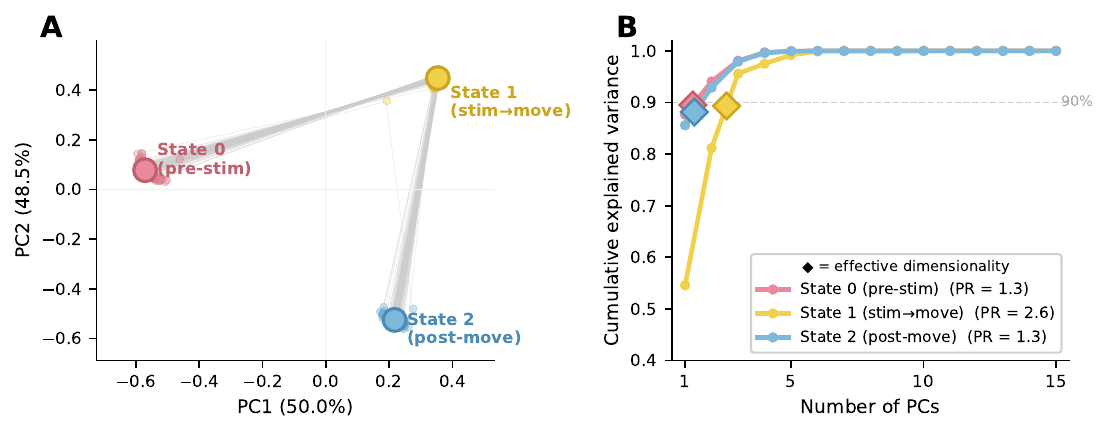}
\caption{\textbf{TWS states occupy orthogonal subspaces with distinct dimensionality.} \textbf{(A)} Session-demeaned PCA of state centroids (243 sessions; pretrained model, first seed). Small dots: per-session centroids; large circles: grand means; gray lines connect the three states within individual sessions. \textbf{(B)} Cumulative explained variance per state. Diamonds mark each state's participation ratio (PR, effective dimensionality). Colors and state labels match Figure~\ref{fig:overview}.}
\label{fig:state_manifold}
\end{figure}

Do the learned TWS states stand for distinct neural regimes, or are they arbitrary partitions of the trial? We extract the state representations of all 243 IBL sessions and analyze their geometry (Figure~\ref{fig:state_manifold}).

\textbf{Orthogonal subspaces.} Session-demeaned PCA of state centroids reveals that the three states occupy near-orthogonal directions in representational space (Figure~\ref{fig:state_manifold}A). The two principal angles between the per-trial Grassmannian bases of the pretrained model (before behavioral finetuning) confirm this: consecutive pairs are nearly orthogonal (S0$\leftrightarrow$S1: $87.0^{\circ}$ and $89.3^{\circ}$, S1$\leftrightarrow$S2: $87.3^{\circ}$ and $89.6^{\circ}$), while the non-adjacent pair shares structure (S0$\leftrightarrow$S2: $9.0^{\circ}$ and $30.6^{\circ}$). Near-orthogonality between consecutive pairs is encouraged by $\Ls_\mathrm{geo}$ (Section~\ref{sec:method}), but the partial structure between non-adjacent states and the per-state dimensionality differences emerge without direct supervision.

\textbf{Dimensionality differs across states.} The participation ratio (PR, effective dimensionality of each state's representation) is $1.3$ before the stimulus, $2.5$ between the stimulus and the movement and $1.3$ after the movement (Figure~\ref{fig:state_manifold}B), so the state that the two boundaries enclose is the one that spreads over more directions. Reaction time follows the same order. Within a session, no principal direction of the pre-stimulus state correlates with reaction time ($|r| \leq 0.03$). The third direction of the stimulus-to-movement state achieves $|r|{=}0.26$ and the third direction of the post-movement state $|r|{=}0.49$. Movement follows the same order over the three states: the largest correlation any direction of a state achieves is $0.08$, $0.26$ and $0.32$. The states that the movement boundary delimits therefore carry the timing, which is what TWS tokens keep across sessions (Section~\ref{sec:main_results}).

\section{Regime Boundary Information Pathway Analysis}
\label{app:gate_decomposition}

\textbf{Why two boundaries.} The IBL protocol records five event times per trial, but only stimulus onset and first movement fall reliably inside the analysis window and mark a clear shift in population dynamics~\citep{churchland2012neural}. Cutting the trial more finely than these two events would also leak the labels: a narrow window placed after the first movement stays empty whenever the mouse does not move, and its emptiness states the movement label.

Regime boundary tokenization raises one concern: the state tokens $\vh$ could pass the boundary time $b_\mathrm{move}$ straight to the decoder and carry no neural content. Reaction time satisfies $\mathrm{RT} \propto b_\mathrm{move} - b_\mathrm{stim}$ with $b_\mathrm{stim}$ constant, so a model that recovers $b_\mathrm{move}$ from $\vh$ predicts RT without decoding anything neural.

We decompose the information in TWS state tokens into two pathways: (i) an \emph{architectural timing prior} arising from the interaction between the gate window shape and generic neural temporal structure, and (ii) \emph{trial-specific neural content} within each epoch.

\textbf{Null result: constant activity carries zero boundary information.} The normalized weights $\bar{w}_k(t) = w_k(t)/\sum_{t'} w_k(t')$ guarantee $\sum_t \bar{w}_k(t) = 1$. When neural activity is temporally constant ($\vx(t) = \vc$ for all $t$), the state mean $\bar{\vx}_k = \sum_t \bar{w}_k(t)\,\vx(t)$ reduces to $\bar{\vx}_k = \vc$ regardless of $b_\mathrm{move}$. The gate cannot transmit boundary information without temporal modulation in the input signal. This rules out direct metadata passthrough.

\textbf{Controlled decomposition.} Real neural activity has temporal structure, such as peri-stimulus modulation and movement-onset transients, and that structure interacts with the position of the gate window. We measure the interaction with a \emph{within-session} ablation on 226 IBL sessions (an earlier version of our session list; the analysis takes only the learned gate offsets of the Table~\ref{tab:main} model and keeps trials whose first movement falls inside the window) ($N{\geq}20$ neurons, ${\geq}100$ valid trials each). For each session, we compute state means $\bar{\vx}_k$ under the four conditions of Table~\ref{tab:gate_decomp} and train a 5-fold cross-validated Ridge regressor (trial split within session) to predict $b_\mathrm{move}$. The evaluation is within-session because the state means $\bar{\vx}_k \in \R^N$ are neuron-indexed and two sessions with different neuron sets cannot be compared in that space. The cross-session values of Table~\ref{tab:main} need the full TWS pipeline, cross-attention into $\R^d$, to make the representations comparable. The four conditions are:

\begin{enumerate}
    \item \textbf{Real} ($b_\mathrm{move}$-aligned gate, matched neural data): standard TWS forward pass.
    \item \textbf{Gate-only} (trial-shuffled $\vx$, real gate): preserves the gate window shape but breaks the correspondence between neural content and $b_\mathrm{move}$. Any remaining $R^2$ reflects temporal fingerprinting: generic temporal modulation (e.g., post-stimulus rate increase common to all trials) interacting with gate position.
    \item \textbf{Neural-only} (shuffled $b_\mathrm{move}$, real $\vx$): applies a mismatched gate to each trial. Any remaining $R^2$ reflects neural temporal structure that encodes movement timing independently of the gate.
    \item \textbf{No-gate baseline}: time-averaged firing rate $\bar{\vx} = T^{-1}\sum_t \vx(t)$ (no windowing).
\end{enumerate}

\begin{table}[!ht]
\caption{Regime boundary timing decomposition (226 IBL sessions, within-session 5-fold CV Ridge on raw state means $\bar{\vx}_k \in \R^{3N}$, mean $\pm$ std across sessions). Not directly comparable to cross-session $R^2$ in Table~\ref{tab:main}.}
\label{tab:gate_decomp}
\begin{center}
\small
\begin{tabular}{lcc}
\toprule
Condition & $R^2$ (predicting $b_\mathrm{move}$) & Interpretation \\
\midrule
Real (aligned gate + neural) & $0.281 \pm 0.192$ & Full signal \\
Gate-only (shuffled $\vx$) & $0.149 \pm 0.105$ & Temporal fingerprint \\
Neural-only (shuffled gate) & $0.274 \pm 0.160$ & Neural temporal content \\
No-gate (time average) & $0.104 \pm 0.116$ & Firing rate baseline \\
\bottomrule
\end{tabular}
\end{center}

\end{table}

\textbf{Results} (Table~\ref{tab:gate_decomp}). The gate-only signal ($R^2{=}0.149$) shows that temporal fingerprinting exists. Temporal modulation common to all trials produces state means that correlate with $b_\mathrm{move}$ once the window moves with it. The neural-only signal ($R^2{=}0.274$) is nearly twice as large and is the larger of the two in 195 of 226 sessions (86\%). Trial-specific neural dynamics, not the geometry of the gate, carry most of the movement timing.

\textbf{Event-time information.} The decomposition above is within-session and restricted to trials with a first movement. To measure how much of each label the event time holds in the cross-session setting of Table~\ref{tab:main}, we fit the same leave-one-session-out Ridge probe on a one-hot encoding of the first-movement bin alone, with no neural data. It scores movement MCC $0.908$ and RT $R^2{=}0.443$, against $0.565$ and $0.402$ for TWS trained from scratch (Table~\ref{tab:main}). Movement is defined by that time: a first movement $0$--$2$\,s after stimulus onset (Table~\ref{tab:glossary}). RT runs from stimulus onset to the IBL response time, which follows the first movement by a median of $0.19$\,s. The boundary, in contrast, is the first-movement time wherever it falls. In $11\%$ of the held-out trials the mouse moves within $0.5$\,s before the stimulus (median $64$\,ms): the label is then $0$ and the boundary falls before the stimulus. The rule ``a boundary after the stimulus means a movement'' alone scores MCC $0.90$, and ``any boundary means a movement'' scores $0.56$. We therefore score the saved embeddings of Table~\ref{tab:main} again without these trials (Table~\ref{tab:prestim}). TWS does not read the label from them: without them it rises from $0.565$ to $0.723$, close to the $0.776$ that the event time then holds, whereas PerceiverIO changes little. The boundary-placement ablation adds that TWS keeps about a third of its movement decoding when the boundary holds no label information (Table~\ref{tab:boundary_ablation}b).

\begin{table}[!ht]
\centering
\caption{Cross-session decoding on the 53 held-out IBL sessions with and without the $11\%$ of trials whose first movement precedes the stimulus (probe of Table~\ref{tab:main}; mean $\pm$ SD over 5 model seeds).}
\label{tab:prestim}
\small
\begin{tabular}{lcccc}
\toprule
 & \multicolumn{2}{c}{All trials} & \multicolumn{2}{c}{Without pre-stimulus movements} \\
\cmidrule(lr){2-3}\cmidrule(lr){4-5}
 & Movement MCC & RT $R^2$ & Movement MCC & RT $R^2$ \\
\midrule
TWS from scratch & $0.565 \pm 0.006$ & $0.402 \pm 0.003$ & $0.723 \pm 0.004$ & $0.418 \pm 0.003$ \\
PerceiverIO & $0.303 \pm 0.009$ & $0.169 \pm 0.003$ & $0.324 \pm 0.018$ & $0.181 \pm 0.004$ \\
\textit{Event-time information} & \textit{$0.908$} & \textit{$0.443$} & \textit{$0.776$} & \textit{$0.462$} \\
\bottomrule
\end{tabular}

\end{table}

\textbf{Every representation given the boundary.} TWS receives the first-movement time as input, and POYO, CEBRA and PerceiverIO do not. We append the one-hot boundary to every representation and refit the probe of Table~\ref{tab:main}, with the features z-scored over trials and no labels involved. With the boundary, every representation decodes movement at the level of the boundary alone, and every representation but POYO does so for RT (Table~\ref{tab:boundary_plus}). Over the 53 held-out sessions, TWS from scratch plus the boundary differs from the boundary alone by $+0.001$ in movement MCC ($p{=}0.23$, paired Wilcoxon test over sessions) and $+0.006$ in RT $R^2$ ($p{=}0.56$), and PerceiverIO with the same backbone plus the boundary differs from the boundary alone by $+0.001$ ($p{=}0.57$) and $+0.007$ ($p{=}0.54$). TWS pretrained + finetuned adds $0.002$ in movement MCC ($p{=}8{\times}10^{-4}$) and nothing in RT. Block is the only target where TWS plus the boundary exceeds the boundary alone by more than $0.01$ ($+0.014$ and $+0.018$), and neither difference is significant ($p{=}0.41$ and $0.21$).

\begin{table}[!ht]
\centering
\caption{Cross-session decoding on the 53 held-out IBL sessions when each representation is given the one-hot first-movement bin (Table~\ref{tab:main} probe with z-scored features; mean $\pm$ SD over 5 model seeds; CEBRA is a single run). PerceiverIO uses the same CNN backbone as TWS.}
\label{tab:boundary_plus}
\small
\resizebox{\textwidth}{!}{
\begin{tabular}{lcccc}
\toprule
Representation + boundary & Movement MCC & RT $R^2$ & Choice MCC & Block MCC \\
\midrule
\textit{Event-time information} & $0.908$ & $0.445$ & $0.032$ & $0.019$ \\
TWS from scratch & $0.910 \pm 0.000$ & $0.450 \pm 0.002$ & $0.029 \pm 0.004$ & $0.033 \pm 0.005$ \\
TWS pretrained + finetuned & $0.911 \pm 0.000$ & $0.450 \pm 0.001$ & $0.031 \pm 0.006$ & $0.037 \pm 0.007$ \\
PerceiverIO & $0.909 \pm 0.001$ & $0.452 \pm 0.002$ & $0.021 \pm 0.006$ & $0.025 \pm 0.005$ \\
POYO & $0.900 \pm 0.008$ & $0.408 \pm 0.015$ & $0.021 \pm 0.007$ & $0.022 \pm 0.005$ \\
CEBRA & $0.909$ & $0.444$ & $0.027$ & $0.015$ \\
\bottomrule
\end{tabular}
}

\end{table}

\textbf{Interpretation.} Segmenting at the first-movement boundary gives TWS tokens a timing prior that any windowed-averaging scheme at that boundary would produce, and within sessions trial-specific neural content adds to it (neural-only $0.274$ against gate-only $0.149$). Methods without a gate keep no timing structure across sessions: POYO and CEBRA both decode reaction time at $R^2{=}0.00$ (Table~\ref{tab:main}). Movement and RT share their defining event with the boundary. Neither target therefore measures neural decoding beyond the boundary. They measure how much regime-aligned timing structure survives the TWS compression.

\section{Backbone Sweep}
\label{app:backbone_sweep}

\begin{figure}[!ht]
\centering
\includegraphics[width=\textwidth,height=0.86\textheight,keepaspectratio]{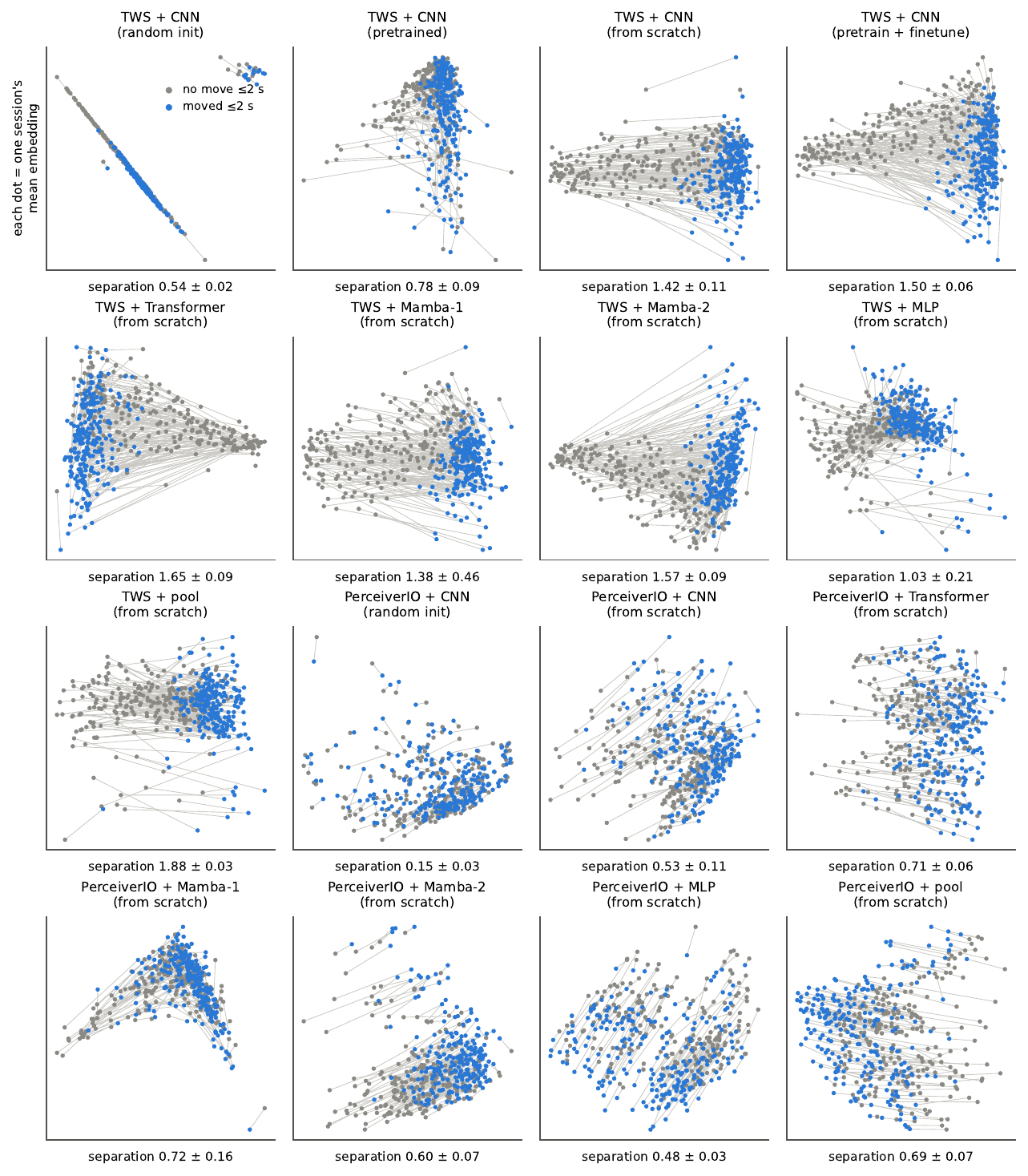}
\caption{Session means of every tokenizer $\times$ backbone $\times$ training combination, drawn as in Figure~\ref{fig:representation}A.}
\label{fig:representation_all}
\end{figure}

\begin{figure}[!ht]
\centering
\includegraphics[width=\textwidth]{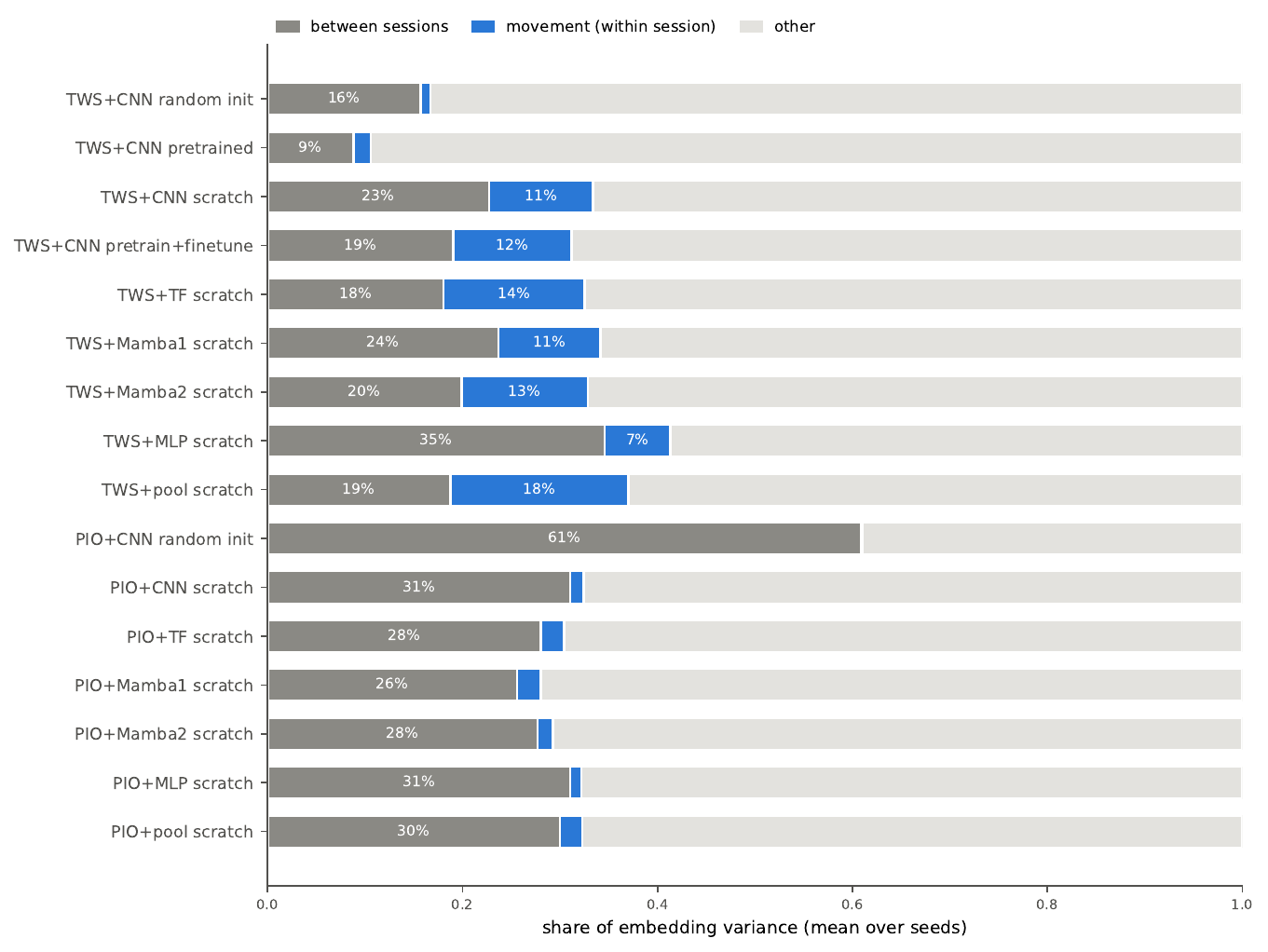}
\caption{What the trial-embedding variance of each combination is made of, drawn as in Figure~\ref{fig:representation}B.}
\label{fig:representation_all_variance}
\end{figure}

\begin{figure}[!ht]
\centering
\includegraphics[width=\textwidth,height=0.86\textheight,keepaspectratio]{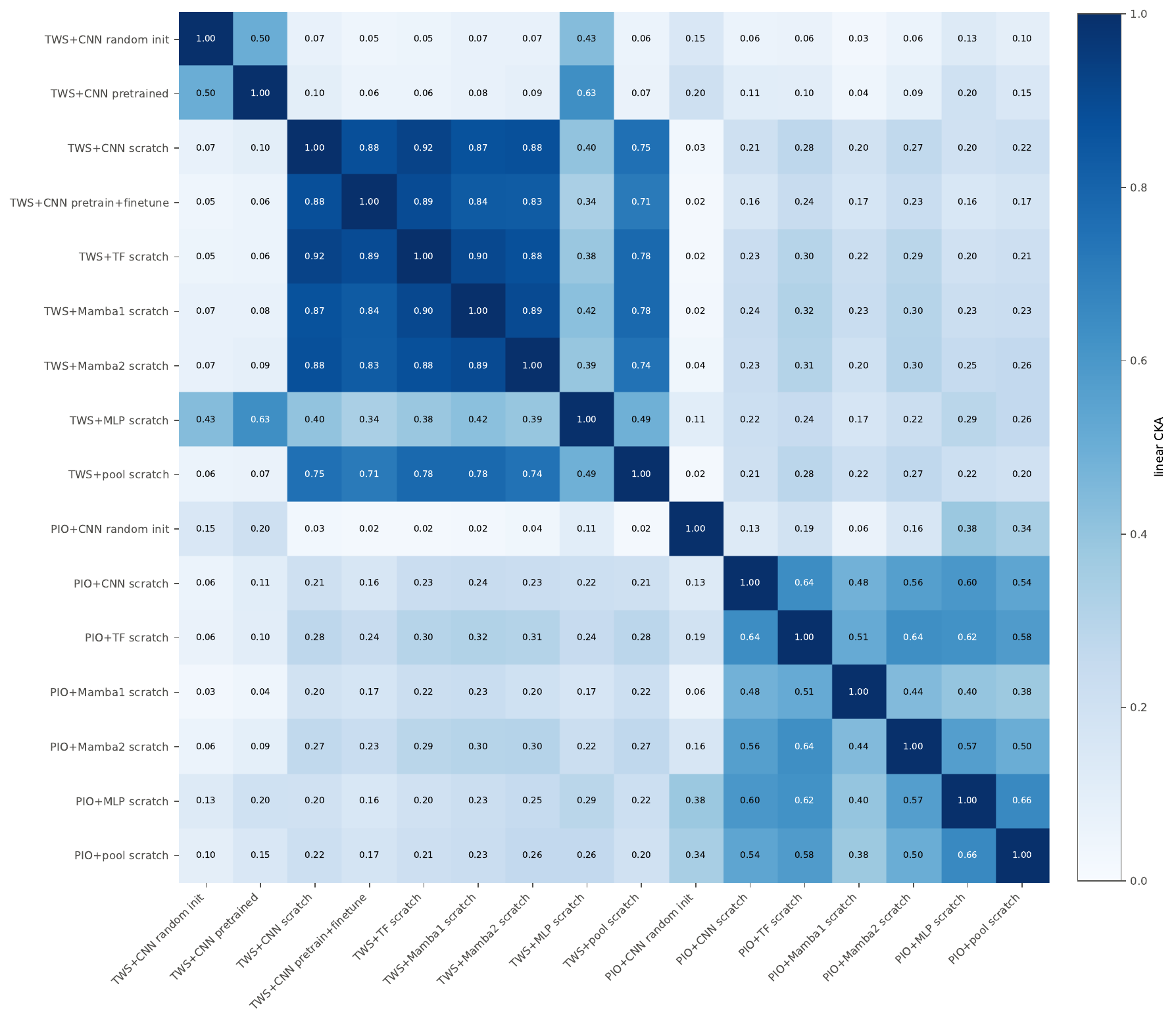}
\caption{Linear CKA between the trial embeddings of every combination (first seed, 20{,}000 shared trials).}
\label{fig:representation_all_cka}
\end{figure}

We replace the backbone while keeping TWS, the pretraining and finetuning recipe, and the evaluation fixed: mean pooling followed by layer normalization (no token mixing), an MLP on the pooled tokens, a CNN (a kernel-3 convolution over the six tokens in each of four blocks), a Transformer, Mamba-1 and Mamba-2. Table~\ref{tab:backbone_ablation} gives cross-session decoding on IBL for each of them. Decoding on IBL is the same within $0.03$ MCC for every backbone that mixes the tokens. Transfer to the other datasets differs between backbones. On Steinmetz and NLB we compare backbones trained on all 243 IBL sessions. From the same TWS tokens, the convolutional backbone achieves Steinmetz choice $0.155 \pm 0.006$ and NLB movement $0.378 \pm 0.008$, and the Mamba backbones land within $0.03$ of it (Mamba-1 over 4 seeds). The Transformer falls behind on both ($0.122 \pm 0.054$ and $0.299 \pm 0.060$), with seed-to-seed spreads an order of magnitude wider. On reach direction, scored with the held-out models, the Transformer ($0.201 \pm 0.006$) stays close to the convolutional backbone ($0.232 \pm 0.010$). Figures~\ref{fig:representation_all}, \ref{fig:representation_all_variance} and \ref{fig:representation_all_cka} show the representation of every tokenizer, backbone and training combination: its session means, the composition of its variance, and how similar the combinations are to one another. Every trained TWS combination separates the two movement states more than every trained PerceiverIO one ($1.03$--$1.88$ against $0.48$--$0.72$), and the tokenizer also decides how similar two embeddings are. Among the four token-mixing backbones trained from scratch, the TWS embeddings agree with one another at a linear CKA of $0.87$--$0.92$ and the PerceiverIO embeddings at $0.44$--$0.64$, while a TWS embedding and a PerceiverIO embedding agree at $0.20$--$0.32$. Backbone and tokenizer trade off in the variance as well: with TWS tokens the convolutional backbone keeps $23\%$ of the variance between sessions and the Transformer $18\%$, against $31\%$ and $28\%$ for the same two backbones on PerceiverIO tokens.

\section{Coding Rate Reduction of the Representations}
\label{app:rate_reduction}

Low-dimensional structure is the property TWS is built to extract, so we also measure it directly with the coding rate of \citet{yu2020learning}. For $n$ trial embeddings normalized to unit length and stacked in $\mZ \in \R^{n \times d}$,
\begin{equation}
R(\mZ) = \tfrac{1}{2}\log\det\!\left(\mI + \tfrac{d}{n\epsilon^2}\mZ^\top\mZ\right), \qquad
\Delta R(\mZ, \Pi) = R(\mZ) - \sum_j \tfrac{n_j}{n}\, R(\mZ_j),
\end{equation}
where $\Pi$ partitions the trials into groups $\mZ_j$ of size $n_j$ and $\epsilon^2 {=} 0.5$. $R$ is the number of nats needed to encode the embeddings up to precision $\epsilon$, a log-volume of the embedding cloud. $\Delta R$ is how much of that volume the grouping explains. It is large when each group is compact and the groups occupy different directions. It is near zero when every group is spread like the whole set. Unlike the variance shares of Figure~\ref{fig:representation}B, which compare group means, $\Delta R$ also counts differences in the directions a group occupies. We group trials by movement (behavioral structure) and by session (session identity), using the same trials for both.

Table~\ref{tab:rate_reduction} shows that learning puts behavior into the structure of the representation, and that the tokenizer decides how much. At random initialization neither tokenizer organizes its volume by behavior: the movement $\Delta R$ is $0.05$ for TWS and $0.03$ for PerceiverIO. Training from scratch raises it to $0.84$ for TWS and to $0.38$ for PerceiverIO. The PerceiverIO cloud is also the larger one ($R{=}56$ against $40$), so behavior explains a smaller share of its volume. Training adds session structure as well: the session $\Delta R$ rises from $3.0$ to $12.7$ for TWS and from $3.0$ to $14.3$ for PerceiverIO. Pretraining moves TWS toward invariance. After pretraining and finetuning the session term falls to $7.0$ and most of the movement term remains ($0.61$). Absolute values depend on $\epsilon$ and on the overall volume $R$, which differs between models, so we compare only large differences. \citet{yu2023whitebox} derive self-attention as a step that compresses tokens against a set of subspaces, which matches TWS tokens being subspace bases. We did not test that link.

\begin{table}[!ht]
\centering
\caption{Coding rate $R$ and its reduction $\Delta R$ (nats) when IBL trial embeddings are grouped by movement or by session (243 sessions, all trials with a movement label; $\epsilon^2{=}0.5$). Models trained without the 53 held-out sessions and with the clipped boundary. Mean $\pm$ SD over 5 model seeds. Other backbones follow with Table~\ref{tab:backbone_ablation}.}
\label{tab:rate_reduction}
\small
\setlength{\tabcolsep}{4pt}
\begin{tabular}{llccc}
\toprule
Tokenizer (backbone) & Training & $R$ & $\Delta R$ (movement) & $\Delta R$ (session) \\
\midrule
\multirow{3}{*}{TWS (CNN)} & random init & $11.65 \pm 1.27$ & $0.049 \pm 0.004$ & $3.00 \pm 0.32$ \\
 & from scratch & $39.71 \pm 6.27$ & $0.836 \pm 0.108$ & $12.67 \pm 2.29$ \\
 & pretrained + finetuned & $28.27 \pm 4.16$ & $0.607 \pm 0.079$ & $6.99 \pm 1.51$ \\
\midrule
PerceiverIO (Transformer) & random init & $10.13 \pm 0.59$ & $0.030 \pm 0.005$ & $3.03 \pm 0.17$ \\
PerceiverIO (Transformer) & from scratch & $56.07 \pm 4.97$ & $0.382 \pm 0.029$ & $14.26 \pm 1.06$ \\
\bottomrule
\end{tabular}

\end{table}

\section{Cross-Dataset Transfer Controls}
\label{app:transfer_controls}

\textbf{Level and pattern of each regime.} TWS summarizes the population of every regime without neuron identity.
Two features of the same three regimes separate the parts of that summary. \emph{Level} is the mean activity over
all recorded neurons. \emph{Pattern} adds the standard deviation and the 10/25/50/75/90\% quantiles over neurons.
Both are invariant to the order of the neurons, and both use the boundaries the model receives. The probe is the
one of Table~\ref{tab:cross_dataset}; Table~\ref{tab:what_transfers} gives the result.
\emph{Level over time} keeps the population mean in every time bin instead of every regime, so it adds the time course inside each regime that TWS averages away.
Over Level, the time course raises NLB movement by a third and doubles reach direction. It raises choice tenfold and lowers feedback.
PerceiverIO keeps this time course in its patches and leads TWS on NLB movement (Table~\ref{tab:xds_pio}); on reach direction TWS leads despite averaging it away (\texttt{scripts/what\_transfers\_time.py}).

\begin{table}[!ht]
\centering
\caption{Cross-dataset MCC of model-free population summaries, against TWS (Table~\ref{tab:cross_dataset}).}
\label{tab:what_transfers}
\small
\begin{tabular}{llcccc}
\toprule
Dataset & Target & Level & Pattern & Level over time & TWS \\
\midrule
\multirow{3}{*}{Steinmetz} & movement & $0.419$ & $0.441$ & $0.475$ & $\mathbf{0.581}$ \\
 & feedback & $0.160$ & $0.174$ & $0.030$ & $\mathbf{0.188}$ \\
 & choice & $0.012$ & $0.027$ & $0.126$ & $\mathbf{0.151}$ \\
\midrule
NLB & movement & $0.354$ & $0.421$ & $\mathbf{0.476}$ & $0.392$ \\
Gallego et al. & reach direction & $0.057$ & $0.083$ & $0.116$ & $\mathbf{0.232}$ \\
\bottomrule
\end{tabular}

\end{table}

\textbf{A second identity-free tokenizer and a random-bias reference.} Table~\ref{tab:xds_pio} adds two references to Table~\ref{tab:cross_dataset}, under the same probe. PerceiverIO with the CNN backbone, trained from scratch on IBL like TWS and then frozen, has no neuron or session embedding either. It transfers as well as TWS: it leads on Steinmetz choice ($p{=}0.018$, paired Wilcoxon test over sessions) and on NLB movement ($p{=}0.006$), TWS leads on reach direction ($p{=}2{\times}10^{-6}$), and the two do not differ on Steinmetz feedback and movement. Untrained, PerceiverIO already decodes choice at $0.169$. TWS (random init) has zero encoder biases. With random biases of the norm of the trained model's biases it reaches three quarters of the trained TWS on choice and on reach direction, and training still adds to choice ($p{=}0.018$), reach direction ($p{<}10^{-10}$) and NLB movement ($p{=}0.012$).

\begin{table}[!ht]
\centering
\caption{Cross-dataset MCC of two frozen tokenizers without neuron or session embeddings, both with the CNN backbone (protocol of Table~\ref{tab:cross_dataset}; mean $\pm$ SD over 5 seeds). The two TWS rows are the runs of Table~\ref{tab:cross_dataset}.}
\label{tab:xds_pio}
\small
\setlength{\tabcolsep}{4pt}
\resizebox{\textwidth}{!}{
\begin{tabular}{lccccc}
\toprule
 & \multicolumn{3}{c}{Steinmetz 2019} & Gallego et al. & NLB MC\_Maze \\
\cmidrule(lr){2-4}\cmidrule(lr){5-5}\cmidrule(lr){6-6}
Method & Choice & Feedback & Movement & Reach direction & Movement \\
\midrule
TWS from scratch & $0.151 \pm 0.015$ & $\mathbf{0.188 \pm 0.008}$ & $0.581 \pm 0.003$ & $\mathbf{0.232 \pm 0.010}$ & $0.392 \pm 0.010$ \\
TWS (random init) & $0.036 \pm 0.016$ & $0.077 \pm 0.017$ & $0.550 \pm 0.007$ & $0.128 \pm 0.002$ & $0.247 \pm 0.010$ \\
TWS (random init, random bias) & $0.114 \pm 0.009$ & $0.178 \pm 0.005$ & $\mathbf{0.592 \pm 0.002}$ & $0.173 \pm 0.004$ & $0.361 \pm 0.005$ \\
PerceiverIO, trained on IBL & $\mathbf{0.190 \pm 0.016}$ & $0.176 \pm 0.004$ & $0.572 \pm 0.004$ & $0.197 \pm 0.013$ & $\mathbf{0.443 \pm 0.020}$ \\
PerceiverIO (random init) & $0.169 \pm 0.024$ & $0.158 \pm 0.014$ & $0.570 \pm 0.009$ & $0.160 \pm 0.009$ & $0.419 \pm 0.017$ \\
\bottomrule
\end{tabular}
}

\end{table}

\textbf{Time-shuffled pretraining.} We permute the 100 time bins of the pretraining data independently for every neuron and trial. The permutation keeps each trial's spike counts and removes temporal order and within-trial co-firing. The pretraining recipe, the backbone and the 53 excluded sessions are those of TWS pretrained, and each model is scored with the probe of Table~\ref{tab:cross_dataset}. Time-shuffled pretraining transfers at least as well as intact pretraining (Table~\ref{tab:time_shuffle}): it is slightly higher on Steinmetz choice and feedback and on reach direction ($p{\leq}0.03$, paired Wilcoxon test over sessions) and equal on both movement targets.

\begin{table}[!ht]
\centering
\caption{Pretrained-only TWS (no behavioral finetuning), cross-dataset MCC, mean $\pm$ SD over 5 seeds.}
\label{tab:time_shuffle}
\small
\begin{tabular}{lccc}
\toprule
Task & Intact pretraining & Time-shuffled pretraining & TWS (random init) \\
\midrule
NLB movement & $0.385 \pm 0.018$ & $0.385 \pm 0.007$ & $0.254 \pm 0.004$ \\
Gallego et al.\ reach direction & $0.215 \pm 0.014$ & $0.232 \pm 0.009$ & $0.126 \pm 0.004$ \\
Steinmetz choice & $0.109 \pm 0.016$ & $0.125 \pm 0.015$ & $0.031 \pm 0.014$ \\
Steinmetz feedback & $0.142 \pm 0.011$ & $0.163 \pm 0.007$ & $0.085 \pm 0.027$ \\
Steinmetz movement & $0.563 \pm 0.003$ & $0.568 \pm 0.005$ & $0.552 \pm 0.004$ \\
\bottomrule
\end{tabular}

\end{table}

\section{Event-Time Information on Cross-Dataset Tasks}
\label{app:boundary_only_xds}

\textbf{Why the boundary decodes IBL but not the other datasets.} The boundary is the first-movement time of the trial. How much it decodes follows from how each target is defined (Table~\ref{tab:boundary_targets}). On IBL movement is a function of that time and RT follows it. On the other datasets movement follows the speed of the wheel or the cursor over the whole trial, which the onset time only constrains, and choice and reach direction do not involve it at all. Each reference uses the probe of its own table: the IBL row the leave-one-session-out Ridge probe of Table~\ref{tab:main}, the cross-dataset row the per-session leave-one-out RidgeClassifierCV of Table~\ref{tab:cross_dataset}.

\begin{table}[!ht]
\centering
\caption{How each target is defined and what the boundary position alone decodes (MCC; RT: $R^2$).}
\label{tab:boundary_targets}
\small
\resizebox{\textwidth}{!}{
\begin{tabular}{lllc}
\toprule
Dataset & Target & Definition & Event-time information \\
\midrule
\multirow{2}{*}{IBL} & movement & a first movement occurs within $2$\,s of stimulus onset & $0.908$ \\
 & RT & stimulus onset to the IBL response time & $0.443$ \\
\midrule
\multirow{3}{*}{Steinmetz} & movement & mean wheel speed of the trial above the session median & $0.319$ \\
 & feedback & reward or no reward & $0.149$ \\
 & choice & wheel turned left, against right or no-go & $-0.009$ \\
\midrule
NLB & movement & mean cursor speed of the trial above the median & $0.260$ \\
Gallego et al. & reach direction & which of the reach targets the trial used & $0.010$ \\
\bottomrule
\end{tabular}
}

\end{table}

On IBL the event that places the first-movement boundary defines movement and RT, so the boundary alone decodes them better than any model (Table~\ref{tab:boundary_ablation}). The other datasets have targets that the boundary does not define, and we ask there whether TWS adds anything beyond it. We fit the cross-dataset probe (per-session leave-one-out, standardized features, RidgeClassifierCV, MCC) on a one-hot encoding of each trial's boundary bin, the boundary TWS itself receives. Steinmetz movement follows wheel speed rather than the boundary event, and Steinmetz choice and Gallego et al.\ reach direction are unrelated to boundary timing by design (Table~\ref{tab:boundary_only_xds}).

\begin{table}[!ht]
\centering
\caption{Cross-dataset MCC of the boundary position alone versus neural representations (TWS: the model of Table~\ref{tab:cross_dataset}, trained without the 53 held-out IBL sessions; mean $\pm$ SD over 5 seeds; PCA and TWS (random init): mean over 5 seeds).}
\label{tab:boundary_only_xds}
\small
\setlength{\tabcolsep}{4pt}
\resizebox{\textwidth}{!}{
\begin{tabular}{llcccc}
\toprule
Dataset & Target & Event-time information & PCA(50) & TWS (random init) & TWS from scratch \\
\midrule
\multirow{3}{*}{Steinmetz} & choice & $-0.009$ & $0.011$ & $0.036$ & $\mathbf{0.151 \pm 0.015}$ \\
 & feedback & $0.149$ & $0.012$ & $0.077$ & $\mathbf{0.188 \pm 0.008}$ \\
 & movement & $0.319$ & $0.390$ & $0.550$ & $\mathbf{0.581 \pm 0.003}$ \\
Gallego et al. & reach direction & $0.010$ & $0.090$ & $0.128$ & $\mathbf{0.232 \pm 0.010}$ \\
NLB & movement & $0.260$ & $0.144$ & $0.247$ & $\mathbf{0.392 \pm 0.010}$ \\
\bottomrule
\end{tabular}
}

\end{table}

\textbf{Choice within movement groups.} Steinmetz choice is correlated with movement, because the fraction of trials with a movement differs between the two classes. A representation that decodes movement could therefore decode choice through it. We decode choice separately within the trials with a movement and within those without, with the probe of Table~\ref{tab:cross_dataset} (Table~\ref{tab:choice_moved}). TWS decodes choice in both groups, while TWS (random init) and PCA stay near chance among the trials without a movement. The boundary position alone decodes choice in neither group. Steinmetz choice is a left turn against a right turn or no turn, so the absence of a movement boundary on no-go trials could also carry it. On the go trials alone, left against right, TWS still leads TWS (random init), PCA and the boundary position.

\begin{table}[!ht]
\centering
\caption{Steinmetz choice MCC within movement groups and on go trials, left against right (cross-session linear probe as in Table~\ref{tab:cross_dataset}; mean $\pm$ SD over 5 seeds; Event-time information as in Table~\ref{tab:boundary_only_xds}).}
\label{tab:choice_moved}
\small
\begin{tabular}{lcccc}
\toprule
Method & All trials & Moved & Did not move & Go trials \\
\midrule
TWS from scratch & $\mathbf{0.151 \pm 0.015}$ & $\mathbf{0.122 \pm 0.010}$ & $\mathbf{0.091 \pm 0.011}$ & $\mathbf{0.105 \pm 0.007}$ \\
TWS (random init) & $0.032 \pm 0.012$ & $0.086 \pm 0.015$ & $0.004 \pm 0.007$ & $0.076 \pm 0.005$ \\
PCA (per session) & $0.011 \pm 0.010$ & $0.054 \pm 0.015$ & $-0.001 \pm 0.003$ & $0.043 \pm 0.009$ \\
\midrule
\textit{Event-time information} & \textit{$-0.009$} & \textit{$0.005$} & \textit{$-0.005$} & \textit{$0.031$} \\
\bottomrule
\end{tabular}

\end{table}

\section{Pretraining Loss Ablation}
\label{app:loss_ablation}

\textit{Backbone note.} The analyses in this section were run with an earlier backbone on the same tokenizer (a multi-scale state-space model) and report accuracy where marked; all rows of the section share that backbone.

\begin{table}[!ht]
\caption{Pretraining loss ablation (IBL, 243 sessions, finetuned). Movement and feedback accuracy.}
\label{tab:loss}
\begin{center}
\small
\begin{tabular}{lcc}
\toprule
Configuration & Movement & Feedback \\
\midrule
Full (MSE + diversity + geodesic + state masking) & 0.873 & 0.817 \\
MSE + diversity (no geodesic) & 0.875 & 0.822 \\
MSE + geodesic (no diversity) & 0.876 & 0.821 \\
MSE only (no geometric losses) & 0.883 & 0.823 \\
No pretraining (from scratch) & 0.889 & 0.824 \\
+ Behavioral supervision during pretraining & 0.887 & 0.825 \\
\bottomrule
\end{tabular}
\end{center}

\end{table}

With 243 sessions every pretraining configuration converges to the same place after finetuning (Table~\ref{tab:loss}). Movement accuracy spans $0.016$ ($0.873$--$0.889$) and feedback $0.008$ ($0.817$--$0.825$). Training from scratch, with no pretraining at all, achieves $0.889$ movement against $0.873$ for the full pipeline. The amount of data decides this convergence. On one of the smaller datasets the IBL checkpoint still adds something: it improves on training from scratch on NLB, and not on Steinmetz or Gallego et al.\ (Appendix~\ref{app:xds_finetune}). With 243 sessions, finetuning finds an effective representation from any initialization, which leaves the geometric prior nothing to add.

\section{Residual Decoding: Orthogonality of Task and Session Information}
\label{app:residual_decoding}

We ask whether movement and session identity occupy orthogonal subspaces. We remove the linear session component from the trial embeddings by subtracting each session's mean embedding, computed from that session's own unlabeled trials. We then decode movement from the residuals with the leave-one-session-out probe of Table~\ref{tab:main} (Table~\ref{tab:residual_decoding}; first model seed).

\begin{table}[!ht]
\centering
\caption{Cross-session movement MCC before (Original) and after (Residual) removing each session's mean embedding. CNN backbone for TWS and PerceiverIO; models trained without the 53 held-out sessions, scored on them (first model seed).}
\label{tab:residual_decoding}
\small
\begin{tabular}{lcc}
\toprule
Model & Original ($\mathbf{h}$) & Residual ($\mathbf{h} - \bar{\mathbf{h}}_s$) \\
\midrule
TWS pretrained + finetuned & 0.555 & 0.561 \\
TWS from scratch & 0.564 & 0.584 \\
TWS (random init) & 0.301 & 0.294 \\
PerceiverIO & 0.310 & 0.324 \\
POYO & 0.000 & 0.001 \\
CEBRA & 0.000 & 0.000 \\
\bottomrule
\end{tabular}

\end{table}

\section{Session-Level Statistics}
\label{app:session_stats}

The $\pm$ in the main tables is the spread over 5 model seeds with the 53 held-out sessions fixed. It says how reproducible a number is, not how consistent it is across sessions. Table~\ref{tab:session_stats} tests the main comparisons over sessions instead. Each held-out session contributes the cross-session score of each representation, averaged over model seeds. We then report the mean paired difference with a $95\%$ bootstrap interval over sessions, a Wilcoxon signed-rank test, and the fraction of sessions where the first representation scores higher. TWS beats PerceiverIO in most sessions. Neither TWS nor PerceiverIO adds to the boundary alone on movement or on reaction time.

\textbf{Seed spread in Figure~\ref{fig:structural_invariance}A.} Over 5 model seeds TWS + CNN from scratch keeps session identifiability $0.489 \pm 0.017$ and decodes cross-session movement at MCC $0.565 \pm 0.006$ on the held-out sessions. At random init it keeps $0.150 \pm 0.004$ and decodes $0.274 \pm 0.018$.

\begin{table}[!ht]
\centering
\caption{Paired comparisons over the 53 held-out IBL sessions (cross-session protocol of Table~\ref{tab:main}; TWS is the from-scratch model with the convolutional backbone unless named). Difference: first minus second, mean over sessions with a bootstrap $95\%$ interval; $p$: Wilcoxon signed-rank; Sessions: fraction of sessions where the first is higher. Each session's score is averaged over 5 model seeds; PerceiverIO uses the same CNN backbone.}
\label{tab:session_stats}
\small
\setlength{\tabcolsep}{3pt}
\resizebox{\textwidth}{!}{
\begin{tabular}{lcccccc}
\toprule
 & \multicolumn{3}{c}{Movement MCC} & \multicolumn{3}{c}{RT $R^2$} \\
\cmidrule(lr){2-4}\cmidrule(lr){5-7}
Comparison & Difference [95\% CI] & $p$ & Sessions & Difference [95\% CI] & $p$ & Sessions \\
\midrule
TWS vs.\ PerceiverIO & $+0.261$ [$+0.210$, $+0.313$] & $3.3\times10^{-9}$ & $91\%$ & $+0.233$ [$+0.170$, $+0.299$] & $2.4\times10^{-8}$ & $91\%$ \\
TWS vs.\ aligned PCA & $+0.216$ [$+0.159$, $+0.273$] & $1.2\times10^{-7}$ & $83\%$ & $+0.176$ [$+0.103$, $+0.254$] & $4.0\times10^{-7}$ & $85\%$ \\
TWS + boundary vs.\ event-time information & $+0.001$ [$-0.000$, $+0.003$] & $0.23$ & $51\%$ & $+0.006$ [$-0.004$, $+0.016$] & $0.56$ & $49\%$ \\
TWS + boundary vs.\ PerceiverIO + boundary & $+0.001$ [$-0.001$, $+0.002$] & $0.35$ & $57\%$ & $-0.001$ [$-0.015$, $+0.014$] & $0.72$ & $51\%$ \\
PerceiverIO + boundary vs.\ event-time information & $+0.001$ [$-0.001$, $+0.003$] & $0.57$ & $49\%$ & $+0.007$ [$-0.010$, $+0.027$] & $0.54$ & $57\%$ \\
\midrule
TWS vs.\ TWS (random init) & $+0.290$ [$+0.250$, $+0.330$] & $4.7\times10^{-10}$ & $98\%$ & $+0.319$ [$+0.245$, $+0.402$] & $6.9\times10^{-9}$ & $92\%$ \\
From scratch vs.\ pretrained + finetuned & $+0.012$ [$-0.005$, $+0.029$] & $0.18$ & $57\%$ & $+0.004$ [$-0.003$, $+0.011$] & $0.02$ & $70\%$ \\
\bottomrule
\end{tabular}
}

\end{table}

\section{Session Directions and Behavior Decoding Directions}
\label{app:session_readout}

We ask whether the directions that separate sessions overlap the directions a linear probe uses to decode
behavior. We work on the z-scored trial embeddings of each model, first model seed. One pooled linear regression
first removes the part of every embedding that movement and reaction time explain, so that sessions differing only
in behavior, a slower mouse for instance, do not define the session directions. The session directions are the top
principal directions of the 243 session means of this residual, as many as explain half of their variance ($k$).
The behavior decoding directions are the weights of Ridge probes for movement and for reaction time, fit on the
190 training sessions.

Table~\ref{tab:session_readout} reports two tests on the 53 held-out sessions. In the first, projecting
out the $k$ session directions leaves movement decoding almost unchanged for every model, $-3\%$ for TWS and for
PerceiverIO, and projecting out $k$ random directions changes nothing. Reaction time overlaps more with the session
directions: TWS loses $37\%$ and PerceiverIO $17\%$. What TWS keeps still exceeds everything PerceiverIO decodes.
In the second test, the two behavior decoding directions identify the session of a trial no better than two random
directions. Session information therefore stays out of the behavior decoding directions, and it is not fully
separable from reaction time. The regression step matters for this conclusion. Without it the session directions
also carry the behavior differences between sessions, and projecting them out costs TWS $18\%$ of movement and
$59\%$ of reaction-time decoding.

\begin{table}[!ht]
\centering
\caption{Session directions (after removing what behavior explains) and behavior decoding directions on the 53 held-out IBL sessions, first model seed. Arrows: cross-session decoding before and after projecting out the $k$ session directions; projecting out $k$ random directions leaves it unchanged. Session identification: nearest-centroid accuracy over the 53 sessions (chance $0.019$).}
\label{tab:session_readout}
\small
\resizebox{\textwidth}{!}{
\begin{tabular}{lccccc}
\toprule
 & & \multicolumn{2}{c}{Remove session directions} & \multicolumn{2}{c}{Session identification from} \\
\cmidrule(lr){3-4}\cmidrule(lr){5-6}
Model & $k$ & Movement MCC & RT $R^2$ & behavior 2D & random 2D \\
\midrule
TWS from scratch & 3 & $0.564 \to 0.546$ & $0.401 \to 0.252$ & $0.050$ & $0.061$ \\
TWS pretrained + finetuned & 3 & $0.550 \to 0.526$ & $0.398 \to 0.251$ & $0.054$ & $0.072$ \\
TWS (random init) & 1 & $0.308 \to 0.302$ & $0.071 \to 0.106$ & $0.080$ & $0.087$ \\
PerceiverIO & 2 & $0.312 \to 0.303$ & $0.170 \to 0.141$ & $0.043$ & $0.065$ \\
\bottomrule
\end{tabular}
}

\end{table}

\textbf{Where session information sits.} We fit the session directions on the 190 training sessions
alone, after the same regression step, and identify the session of each trial of the 53 held-out sessions from
different subsets of dimensions (Table~\ref{tab:session_location}). In the trained models the top session directions carry more session information than as many random directions, and removing them does not lower session identification. Any two random directions identify the session above chance. Session information is therefore spread over the whole
embedding rather than held in a few directions. This holds for TWS and for PerceiverIO, which differ in how much of
it they keep. The two behavior decoding directions carry no more of it than two random directions.

\begin{table}[!ht]
\centering
\caption{Nearest-centroid session identification on the 53 held-out IBL sessions from subsets of embedding dimensions (chance $0.019$; two random directions give $0.05$--$0.10$). First model seed.}
\label{tab:session_location}
\small
\resizebox{\textwidth}{!}{
\begin{tabular}{lcccccc}
\toprule
 & & \multicolumn{4}{c}{Session directions (90\% of the variance of session means)} & \\
\cmidrule(lr){3-6}
Model & All & $k$ & Only these & All but these & $k$ random & \shortstack{Behavior\\directions (2D)} \\
\midrule
TWS from scratch & $0.232$ & 8 & $0.187$ & $0.248$ & $0.145$ & $0.050$ \\
TWS pretrained + finetuned & $0.212$ & 7 & $0.185$ & $0.200$ & $0.132$ & $0.054$ \\
TWS (random init) & $0.126$ & 3 & $0.102$ & $0.098$ & $0.107$ & $0.080$ \\
PerceiverIO & $0.417$ & 12 & $0.332$ & $0.425$ & $0.220$ & $0.043$ \\
\bottomrule
\end{tabular}
}

\end{table}

\section{Pretraining Information Decomposition}
\label{app:info_decomp}

We measure how much nuisance information each embedding keeps at three levels: lab, mouse and session (Table~\ref{tab:info_decomp}). For each level we train a 5-fold cross-validated logistic regression probe ($\ell_1$ penalty, saga solver) on 50K subsampled trial embeddings. MI is $H(Y) - \mathrm{CE}_\mathrm{probe}$, with $H(Y)$ the entropy of the uniform label distribution at that level: 12 labs, 109 mice, 243 sessions.

\begin{table}[!ht]
\centering
\caption{Nuisance information retention: MI$(\mathbf{h};\,Y)/H(Y)$ at each hierarchical level. Lower values indicate greater invariance to the corresponding factor. TWS and PerceiverIO are trained without the 53 held-out sessions, POYO pretrained on all sessions and CEBRA fits one model per session; PerceiverIO uses the same CNN backbone as TWS.}
\label{tab:info_decomp}
\small
\resizebox{\textwidth}{!}{
\begin{tabular}{lcccccc}
\toprule
Factor & TWS pretrained + finetuned & TWS from scratch & TWS (random init) & PerceiverIO & POYO & CEBRA \\
\midrule
Lab      & 0.174 & 0.209 & 0.078 & 0.281 & 0.426 & 0.024 \\
Mouse    & 0.334 & 0.361 & 0.111 & 0.570 & 0.809 & 0.113 \\
Session  & 0.420 & 0.440 & 0.137 & 0.733 & 0.959 & 0.220 \\
\bottomrule
\end{tabular}
}

\end{table}

\section{Site Invariance Across Recording Labs}
\label{app:site_invariance}

\begin{figure}[!ht]
\centering
\includegraphics[width=\textwidth]{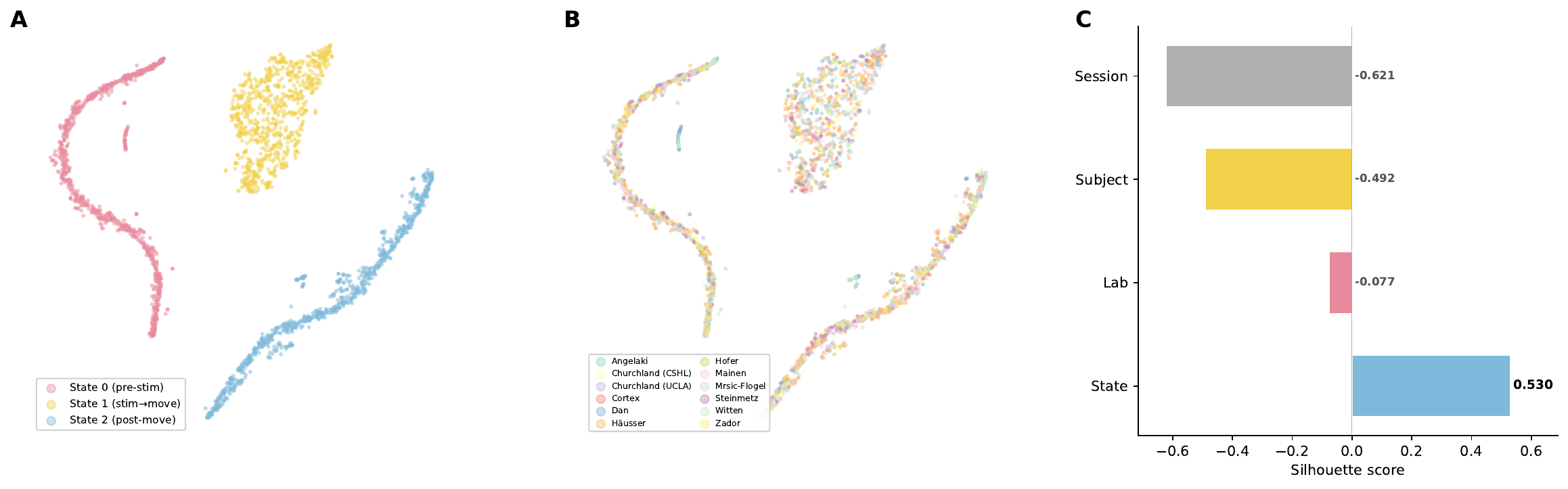}
\caption{State tokens of TWS pretrained (the 53 held-out sessions excluded from training) for all 243 IBL sessions. \textbf{(A)} t-SNE of 2{,}400 state tokens (800 per state) colored by state. \textbf{(B)} The same coordinates colored by recording lab. \textbf{(C)} Silhouette scores of the t-SNE coordinates for state, lab, subject and session.}
\label{fig:site_invariance}
\end{figure}

The IBL Brain Wide Map spans 12 laboratories in 4 countries, each with its own Neuropixels rigs. Recording site is a common source of batch effects in multi-site studies. We test whether TWS state tokens cluster by recording lab or by computational state (Figure~\ref{fig:site_invariance}). In the t-SNE map the tokens group by state (silhouette $0.53$) and not by lab ($-0.08$), subject ($-0.49$) or session ($-0.62$). The session information that a linear probe still finds sits in directions that behavior decoding does not use (Section~\ref{sec:invariance}).

\section{Trial Embeddings of Held-Out Sessions}
\label{app:tsne}

\begin{figure}[!ht]
\centering
\includegraphics[width=\textwidth]{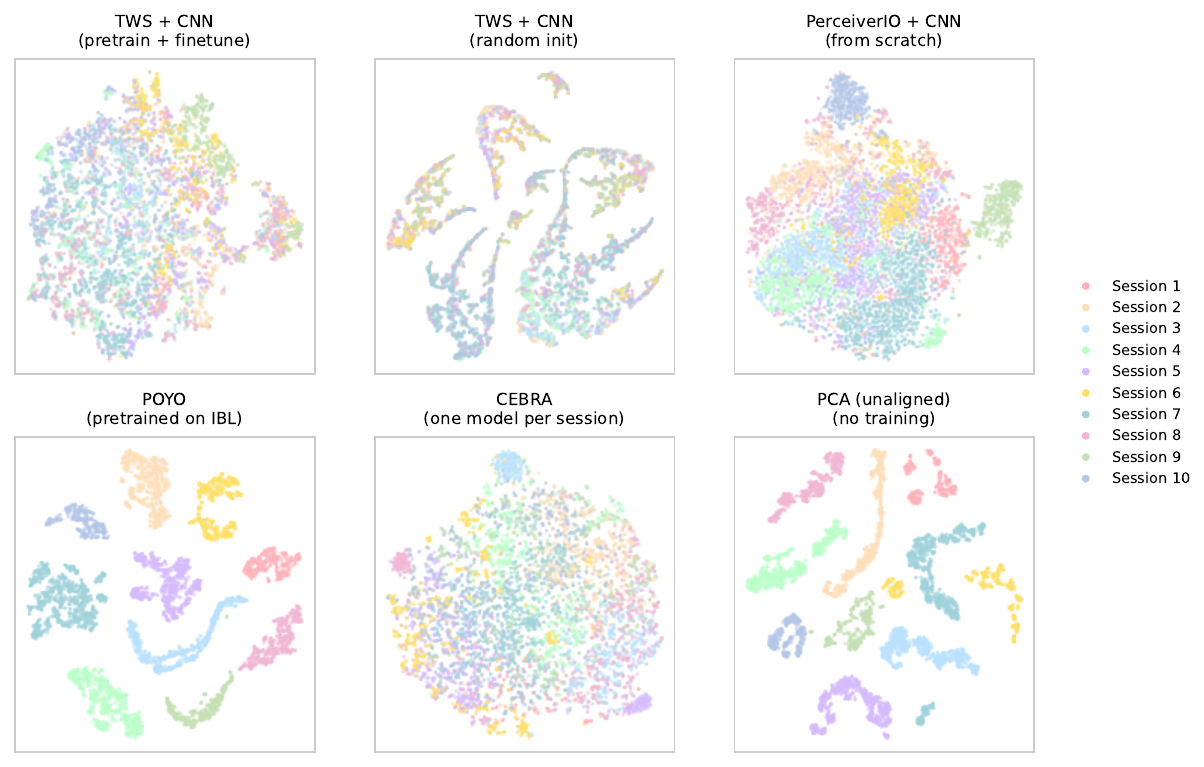}\\[4pt]
\includegraphics[width=0.32\textwidth]{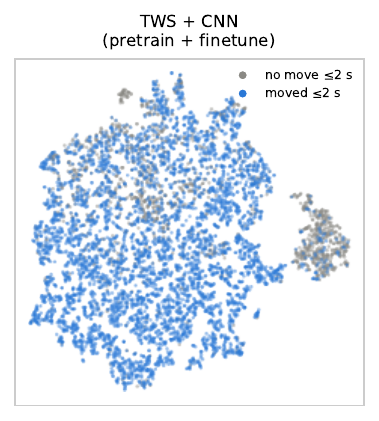}\hfill
\includegraphics[width=0.32\textwidth]{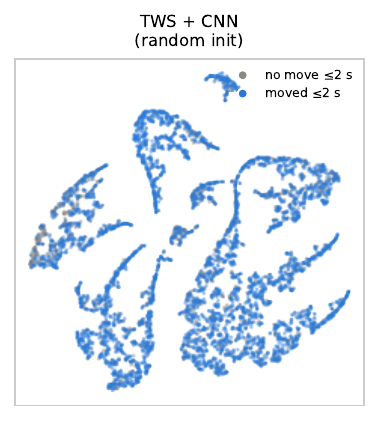}\hfill
\includegraphics[width=0.32\textwidth]{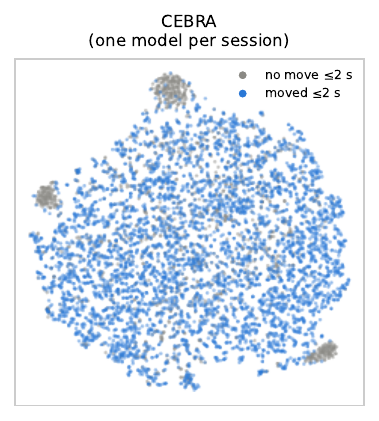}
\caption{2D t-SNE of the trial embeddings of 10 held-out IBL sessions. Top: one color per session. Bottom: the same maps of TWS pretrained + finetuned, TWS (random init) and CEBRA, colored by whether the mouse moved within 2\,s. POYO and PCA separate the sessions, while TWS mixes them and still separates the two behaviors. Nearby trials from other sessions share the movement label above chance for TWS pretrained + finetuned ($+0.13$), but hardly for TWS (random init) ($+0.03$) or CEBRA ($+0.02$), whose separate per-session models overlap only by chance (15 nearest neighbors from other sessions).}
\label{fig:tsne}
\end{figure}

Figure~\ref{fig:tsne} maps the trial embeddings of 10 held-out sessions in two dimensions, for every method
of Table~\ref{tab:main} that produces trial embeddings.

\section{Cross-Dataset Replication: Steinmetz 2019}
\label{app:steinmetz_mi}

\begin{figure}[!ht]
\centering
\includegraphics[width=0.5\textwidth]{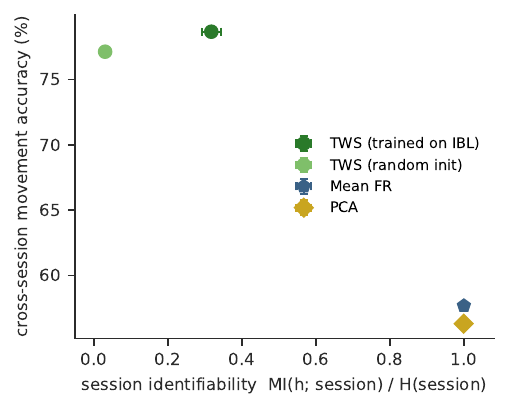}
\caption{Session identifiability (normalized MI of a 39-class probe) against cross-session movement accuracy on the 39 Steinmetz sessions. Mean $\pm$ SD over 5 model seeds.}
\label{fig:steinmetz_mi}
\end{figure}

We repeat the session identifiability analysis of Figure~\ref{fig:structural_invariance}A on Steinmetz 2019 with the frozen TWS of Table~\ref{tab:cross_dataset}, which never saw this dataset (Figure~\ref{fig:steinmetz_mi}). Per-neuron features identify every session and decode movement close to chance ($50\%$, a median split of wheel speed). TWS trained on IBL keeps a third of that session information and decodes movement best; TWS (random init) retains almost none and decodes movement nearly as well, as in Table~\ref{tab:cross_dataset}.

\section{A Shared Decoding Direction Explains Data Efficiency}
\label{app:behavior_axis}

For every model and each of the 53 held-out sessions we z-score the trial embeddings and take one
decoding direction of the session. For movement it is the mean embedding of the trials with a first movement
within 2\,s minus the mean of the trials without. For reaction time it is the covariance of each embedding
dimension with reaction time, z-scored within the session. \emph{Direction consistency} is the cosine between the
unit decoding direction of a session and the mean direction of the other held-out sessions. The \emph{five-session
score} is the movement MCC or the reaction-time $R^2$ of a Ridge probe fit on 5 sessions drawn from the 190
training sessions, with the regularization chosen on a sixth session and the score averaged over 20 draws; $R^2$ is
clipped to $[-1,1]$ per session, as in Table~\ref{tab:main}. \emph{Session identification} is the accuracy of a
nearest-centroid classifier over the 53 held-out sessions, 5-fold within each session. Table~\ref{tab:behavior_axis}
gives all three quantities.

Across the six models the five-session score follows direction consistency for both targets (Spearman
$\rho{=}0.94$ for each). It does not follow session identification ($\rho{=}{-}0.09$ for movement, $0.03$ for
reaction time). The same relation holds inside every model: sessions whose decoding direction agrees with the
others also score higher on five sessions ($\rho$ from $0.58$ to $0.80$, all $p{<}10^{-5}$). Part of that
within-model relation can come from sessions with a stronger behavior signal, which have both a steadier decoding
direction and better decoding, so we rest the claim on the comparison across models.

\begin{table}[!ht]
\centering
\caption{Direction consistency, five-session decoding and session identification on the 53 held-out IBL sessions (mean over 5 model seeds). PerceiverIO uses the same CNN backbone as TWS.}
\label{tab:behavior_axis}
\small
\resizebox{\textwidth}{!}{
\begin{tabular}{lccccc}
\toprule
 & \multicolumn{2}{c}{Movement} & \multicolumn{2}{c}{Reaction time} & \\
\cmidrule(lr){2-3}\cmidrule(lr){4-5}
Model & Consistency & Five-session MCC & Consistency & Five-session $R^2$ & \shortstack{Session\\identification} \\
\midrule
TWS from scratch & $0.82$ & $0.513$ & $0.93$ & $0.313$ & $0.272$ \\
TWS pretrained + finetuned & $0.82$ & $0.507$ & $0.91$ & $0.330$ & $0.228$ \\
TWS pretrained & $0.55$ & $0.383$ & $0.56$ & $0.009$ & $0.132$ \\
TWS (random init) & $0.50$ & $0.182$ & $0.62$ & $0.021$ & $0.123$ \\
PerceiverIO & $0.64$ & $0.254$ & $0.73$ & $0.073$ & $0.406$ \\
PerceiverIO (random init) & $0.40$ & $0.138$ & $0.40$ & $-0.027$ & $0.347$ \\
\bottomrule
\end{tabular}
}

\end{table}

\end{document}